\documentclass[letterpaper, 10 pt, journal, twoside]{IEEEtran}
\IEEEoverridecommandlockouts

\usepackage{cite}
\usepackage{amsmath,amssymb,amsfonts}
\usepackage{algorithmic}
\usepackage{graphicx}
\graphicspath{{./images/}}
\DeclareGraphicsExtensions{.png, .pdf}
\usepackage{textcomp}
\usepackage{xcolor}
\usepackage{url}

\usepackage{color}
\definecolor{best}{rgb}{.2,.6,.2} 
\definecolor{good}{rgb}{.5,.99,.5} 
\definecolor{same}{rgb}{0.1,0.1,0.1}  
\definecolor{bad}{rgb}{.9,.7,.1} 
\definecolor{superbad}{rgb}{.9,.1,.1}   
\definecolor{notworking}{rgb}{.5,.5,.5}

\definecolor{supergood}{rgb}{.2,.6,.2} 
\definecolor{equal}{rgb}{0.1,0.1,0.1}

\usepackage{dsfont}
\usepackage{microtype}

\usepackage{mathptmx}
\usepackage{helvet}

\usepackage[caption=false, font=footnotesize]{subfig}
\usepackage[capitalise]{cleveref}

\usepackage{enumitem}

\DeclareMathOperator*{\argmin}{arg\,min}
\DeclareMathOperator*{\argmax}{arg\,max}

\begin{document}

\title{ \LARGE \bf
Applications of Spiking Neural Networks in Visual Place Recognition
}

\author{Somayeh Hussaini, \IEEEmembership{Member, IEEE}, Michael Milford, \IEEEmembership{Senior Member, IEEE}, and Tobias Fischer, \IEEEmembership{Senior Member, IEEE}
\thanks{
Received 31 July 2024; accepted 1 October 2024. The work of M. Milford was supported in part by the Australian Government under Grant AUS-MURIB000001 associated with ONR MURI Grant N00014-19-1-2571, in part by ARC Laureate Fellowship FL210100156, and in part by Intel Labs. The work
of T. Fischer was supported by Intel Labs. This article was recommended for publication by Associate Editor A. Nuechter and Editor J. Civera upon evaluation of the reviewers’ comments. \textit{(Corresponding Author: Somayeh Hussaini.)}

The authors are with the QUT Centre for Robotics, School of Electrical Engineering and Robotics, Queensland University of Technology, Brisbane, QLD 4000, Australia (e-mail: s.hussaini@qut.edu.au; michael.milford@qut.edu.au; tobias.fischer@qut.edu.au).

Digital Object Identifier 10.1109/TRO.2024.3508053
}
}

\markboth{IEEE Transactions on Robotics. Preprint Version. Accepted October, 2024}{Hussaini \MakeLowercase{\textit{et al.}}: Applications of Spiking Neural Networks in Visual Place Recognition}

\IEEEtitleabstractindextext{%
\begin{abstract}

In robotics, Spiking Neural Networks (SNNs) are increasingly recognized for their largely-unrealized potential energy efficiency and low latency particularly when implemented on neuromorphic hardware. Our paper highlights three advancements for SNNs in Visual Place Recognition (VPR). Firstly, we propose Modular SNNs, where each SNN represents a set of non-overlapping geographically distinct places, enabling scalable networks for large environments. Secondly, we present Ensembles of Modular SNNs, where multiple networks represent the same place, significantly enhancing accuracy compared to single-network models. 
Each of our Modular SNN modules is compact, comprising only 1500 neurons and 474k synapses, making them ideally suited for ensembling due to their small size. 
Lastly, we investigate the role of sequence matching in SNN-based VPR, a technique where consecutive images are used to refine place recognition. 
We demonstrate competitive performance of our method on a range of datasets, including higher responsiveness to ensembling compared to conventional VPR techniques and higher R@1 improvements with sequence matching than VPR techniques with comparable baseline performance. 
Our contributions highlight the viability of SNNs for VPR, offering scalable and robust solutions, and paving the way for their application in various energy-sensitive robotic tasks.

\end{abstract}

}

\maketitle
\IEEEdisplaynontitleabstractindextext

\section{Introduction}
\label{introduction}

\IEEEPARstart{S}{piking} Neural Networks (SNNs) represent a cutting-edge paradigm in neuromorphic computing, mirroring the intricate workings of biological neural systems~\cite{ghosh2009spiking, sandamirskaya2022neuromorphic, schuman2022opportunities, yamazaki2022spiking}. In these networks, every neuron possesses its distinct activation state. Unlike conventional neural networks, where neuron activations are typically continuous values, neurons in SNNs convey information through intermittent spikes, which are initiated when the neuron's activation surpasses a particular threshold~\cite{ghosh2009spiking, gerstner2014neuronal, nunes2022spiking}. These spiking networks exhibit promising attributes when deployed on neuromorphic hardware, offering notable energy efficiency and low-latency data processing~\cite{frady2020neuromorphic,davies2021advancing, sandamirskaya2022neuromorphic, pei2019towards, furber2014spinnaker}. 
Despite these potential advantages, SNNs have seen minimal adaptation in robotics, due to limitations such as the difficulty of supervised training caused by the non-differentiable activation function of spiking neurons
and a lack of tools and resources~\cite{sandamirskaya2022neuromorphic, yik2023neurobench, nunes2022spiking, eshraghian2023training, yamazaki2022spiking}.

One robotics application that could benefit substantially from the emerging neuromorphic computing paradigm is the Visual Place Recognition (VPR) task, a vital process in robotic navigation. 
At its core, the objective is seemingly straightforward: given a query image of a place, find the corresponding place out of a potentially very large list of previously visited places, also called the reference dataset~\cite{schubert2023visual, garg2021your, Lowry2015, tsintotas2022visual, masone2021survey, zhang2021visual}. 
However, there are immense underlying challenges such as changes in appearance due to different times of the day, variations in seasons, weather conditions, and perceptual aliasing (where two geographically distant places may look very similar), leading to significant appearance discrepancies between reference and query images of the same place~\cite{garg2021your, Lowry2015, masone2021survey}.

VPR is a critical component in robot localization tasks such as loop closure detection in Simultaneous Localization and Mapping (SLAM), and global re-localization of mobile robots~\cite{cadena2016past, schubert2023visual, zhang2021visual, tsintotas2022visual}. 
It is also relevant to image retrieval and landmark recognition tasks~\cite{Lowry2015, schubert2023visual}.  
Within robot navigation, VPR can minimize localization errors by recognizing previously visited places and updating the map of the environment despite appearance changes~\cite{Lowry2015, schubert2023visual, tsintotas2022revisiting}, which enables mobile robots to operate over extended periods.

To operate in real-time on resource-constrained robots, such as space exploration and disaster recovery where long mission times are desirable, conventional state-of-the-art VPR methods might not be applicable as they often have high computational demands~\cite{frenkel2023bottom}, motivating the use of SNNs within VPR. 
We are further inspired by the remarkable ability of animals with relatively small brains, such as rodents, to effectively perform navigation in complex environments. 
Although SNNs are not as mature as their deep learning counterparts, the benefits of SNNs include low power usage and low latency, particularly when deployed on neuromorphic hardware~\cite{davies2021advancing,sandamirskaya2022neuromorphic,schuman2022opportunities}. VPR is a particularly intriguing task for SNNs as VPR is amenable to enhancement strategies such as ensembling and using temporal data via sequences of images.

The objective of our work is to explore an alternative SNN-based approach to current state-of-the-art VPR techniques based on deep learning that is scalable and efficient, and has the prospect to be ideally suited for energy-sensitive robotic tasks. %
We achieve this by demonstrating the potential of a simple three-layer SNN in achieving place recognition at a significant scale.
Our approach leverages widely adopted strategies for enhancing the robustness of conventional machine learning approaches, specifically focusing on modularity, ensembling, and sequence matching techniques.
The subsequent paragraphs will offer a succinct overview of these key strategies.

Modularity, a prominent concept in machine learning, entails the design of systems comprising distinct modules, each dedicated to a specific task~\cite{auda1999modular, rauker2023toward, amer2019review}. These modules can be combined to form more intricate systems, offering scalability benefits beyond what individual modules can achieve~\cite{auda1999modular, rauker2023toward, amer2019review}. 
Modularity has been employed in
\cite{colosi2020plug} for a SLAM system with heterogeneous sensor configuration, 
a 3D place recognition task~\cite{dube2017segmatch},
a condition and environment-invariant place recognition task~\cite{garg2017improving}, 
and a SLAM system~\cite{blanco2019modular}. These works assign a module to each structurally different sub-task of a system. In our work, we similarly use the concept of modularity to enable robustness to more challenging scenarios, however we create modules with the same architecture which are assigned to learn small segments of the dataset.

Ensembling\footnote{We note ensembling~\cite{ganaie2022ensemble} and model fusion~\cite{li2023deep} have been used interchangeably in literature to denote the practice of combining multiple models or feature representations for improved performance. Throughout this work, we refer to this technique as ensembling.}
is the strategy of combining multiple models to boost accuracy, reduce overfitting, and yield more robust models~\cite{ghosh2009spiking, ganaie2022ensemble, yang2023survey, dietterich2000ensemble, zhou2012ensemble, sagi2018ensemble}. 
While there is an inherent trade-off between the number of ensemble members and the energy efficiency of a system~\cite{ganaie2022ensemble, yang2023survey}, the benefits of ensembling, particularly its enhanced accuracy and reliability, are invaluable to navigation-related robotics applications. 
Ensembling has been previously utilized in the context of
monocular SLAM~\cite{wu2020eao} for improved data association, 
terrain segmentation for learning data collected at different times~\cite{procopio2009learning}, 
and place recognition with image-based~\cite{arcanjo2023music, malone2023boosting} and event-based~\cite{fischer2020event} input, 
where multiple VPR techniques fuse to enhance place recognition robustness. 
Similarly, our work also uses ensembling to improve upon the robustness and generalization ability of our system but differs with these works in terms of how we define our ensemble members. Our ensemble members all are tasked to do the same task, have the same architecture and differ in random shuffling of input images and initial learned weights. A similar approach of employing ensembles was previously demonstrated in uncertainty estimation to provide sufficient diversity among the members, and improve the overall performance of the system~\cite{lakshminarayanan2017simple}.

In the context of VPR, a sequence matching technique uses consecutive reference frames instead of single frames, to match a query image of a place to its corresponding reference image~\cite{milford2012seqslam, garg2021seqnet, garg2022seqmatchnet}. By analyzing a series of images, this technique enables enhanced resilience against temporary environmental disruptions and improves localization accuracy~\cite{milford2012seqslam, garg2021seqnet, garg2022seqmatchnet}. 
Sequence matching techniques have been widely employed in place recognition~\cite{milford2012seqslam, schubert2021fast, garg2022seqmatchnet, mereu2022learning, garg2021seqnet, facil2019condition, naseer2015robust}, where often a decoupled approach consisting of an initial image-based retrieval method and subsequent sequence score aggregation~\cite{milford2012seqslam, schubert2021fast, naseer2015robust} is employed. 
Similar to these techniques, our paper uses sequence matching as an additional step after image retrieval. We further demonstrate the effectiveness of sequence matching for SNN-based approaches, and provide an indicator for responsiveness of conventional and SNN-based VPR methods to sequence matching.

\begin{figure*}[t]
\centering
\includegraphics[width=\linewidth]{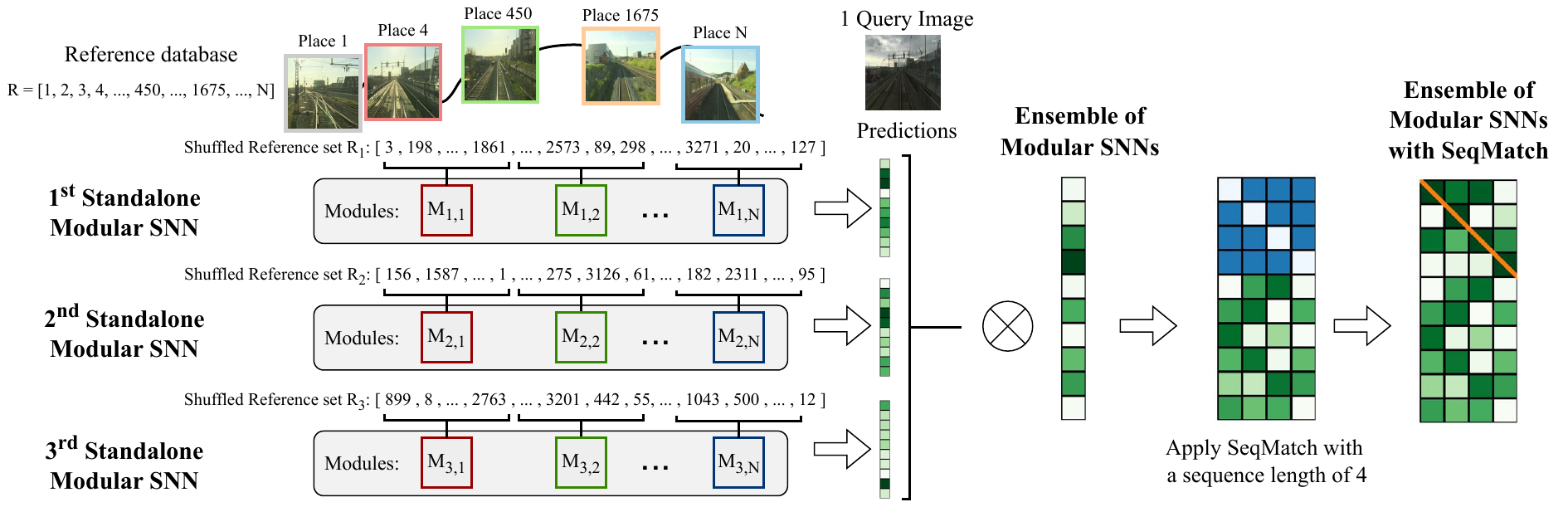}

\caption{\textbf{Schematic of the proposed algorithm:} 
The basic building blocks in our work are independent Spiking Neural Network (SNN) modules that learn small subsets of the reference database. At inference time, the place predictions of all these modules are fused in parallel in what we dub a ``Standalone Modular SNN'', enabling the scalability of our approach to a large number of places.  
We further make use of the potential massively parallel processing capabilities of neuromorphic processors by introducing ensembles in which multiple Modular SNNs learn representations for the same place, and demonstrate that SNNs are more responsive to ensembling compared to conventional techniques. Finally, we demonstrate the high responsiveness of these Ensembles of Modular SNNs to sequence matching.
}
\label{fig:frontpage_v2}
\vspace*{-0.4cm}
\end{figure*}

\noindent In this work, we claim the following contributions (\Cref{fig:frontpage_v2}): 

\begin{enumerate}
    \item We present the concept of modular spiking neural networks (Modular SNNs) for scalable place recognition. Each module of the network \textit{specializes} in a small subset of places in the environment at training time and operates \textit{independently} of all other networks.
    After training the modules independently, we address the lack of \emph{global} regularization by detecting hyperactive neurons, those that frequently respond to images \emph{outside} their training data, and subsequently ignoring them during deployment. The query image at inference time is provided to all modules in parallel, and their place predictions are fused.

    \item While the first contribution serves as a functional framework for scalable place recognition with Modular SNNs, this second contribution further enhances its capabilities through Ensembles of Modular SNNs, where \emph{multiple networks} learn a representation for the \emph{same place}, leading to improved robustness and generalization capabilities. Each ensemble member constitutes a Modular SNN with variations in the weight initialization and the set of distinct places learned by a module. Our results demonstrate that SNN ensemble members exhibit higher variations in their match predictions compared to conventional counterparts, which results in significantly higher responsiveness to ensembling.

    \item We analyze the responsiveness of our Modular and Ensemble of Modular SNNs to sequence matching, which captures the temporal information inherent in the data, by considering multiple consecutive reference places for predicting a single query image, as opposed to considering one single reference image. 
    We also present an indicator that predicts the effectiveness of applying sequence matching to both conventional VPR methods and our spiking networks, to provide insights into the improvements conferred by applying a sequence matching technique. 
    
    \item We provide extensive evaluations of our SNN performance, and compare them to conventional VPR techniques, i.e.~Sum-of-absolute differences (SAD)~\cite{milford2012seqslam}, \mbox{DenseVLAD}~\cite{torii201524}, NetVLAD~\cite{Arandjelovic2018}, AP-GeM~\cite{revaud2019learning}, GCL~\cite{leyva2023data}, CosPlace~\cite{berton2022rethinking}, and MixVPR~\cite{ali2023mixvpr}
    across multiple benchmark datasets, namely Nordland~\cite{sunderhauf2013we}, Oxford RobotCar~\cite{RobotCar}, SFU-Mountain~\cite{bruce2015sfu}, Synthia Night to Fall~\cite{ros2016synthia}, and St Lucia~\cite{milford2008mapping}. 
    Compared to previous work~\cite{Hussaini2022}, we evaluate our approach on datasets that are up to two orders of magnitude larger. 
    We also introspect our SNNs and provide insights to their responsiveness when paired with sequence matching, and contrast that to that of conventional VPR techniques. 
    Finally, we provide a proof-of-concept deployment of our Modular SNN on AgileX's Scout Mini robot~\cite{agilex_scout_mini} in a small indoor environment, operating in real-time on CPU.  

\end{enumerate}

The first contribution on Modular SNNs was previously presented at the IEEE International Conference on Robotics and Automation (ICRA) 2023~\cite{hussaini2023ensembles}. 
This work substantially extends on~\cite{hussaini2023ensembles} by employing ensembling and sequence matching techniques to significantly enhance place recognition capabilities and analyzing the impact of these techniques across the entire set of evaluated models. 
We introduce an indicator that predicts the responsiveness of both conventional VPR techniques and our spiking networks to sequence matching. 
We also provide significantly extended evaluations, now covering six datasets, up from the initial two.
Furthermore, we demonstrate the scalability of our approach by applying it to datasets that are up to an order of magnitude larger in terms of the number of learned places, and we do not require a calibration step anymore, which previously required paired images for a subset of the query/reference datasets.
For the first time, we demonstrate our Modular SNN in a real-time proof-of-concept CPU-based robot deployment in a small indoor environment. 
We also benchmark against seven distinct VPR techniques, encompassing 14 variations, compared to the three techniques previously compared against in~\cite{hussaini2023ensembles}. 
Our new contributions demonstrate the enhanced effectiveness of SNNs for VPR, including increased scalability and robustness, paving the way for application in energy-efficient robotic navigation tasks.

The rest of this article is organized as follows. In \Cref{related_works},
we will delve into the related works to provide context for SNNs
and VPR; in \Cref{methodology}, we will discuss our methodology;
experimental setup is given in \Cref{exp_setup}; \Cref{res:I} presents
our results with analysis; and finally, \Cref{discussions_conclusions} concludes this article.

\section{Related works}
\label{related_works}

In this section, we offer an overview of neuromorphic computing and Spiking Neural Networks (SNNs) (\Cref{RW:neuromorphic_n_SNN}) and explore the applications of spiking neural networks in robot localization (\Cref{RW:snns_for_loc}). We then delve into the Visual Place Recognition (VPR) task (\Cref{RW:vpr}), non-spiking biologically inspired VPR approaches (\Cref{RW:bbio-inspired_vpr}) and provide insights into ensembling (\Cref{RW:ensembling}) and sequence matching techniques (\Cref{RW:seq_matching}).

\subsection{Neuromorphic Computing and SNNs}
\label{RW:neuromorphic_n_SNN}

The field of neuromorphic computing focuses on hardware, sensors, and algorithms inspired by biological neural networks, aiming to capture the robustness, generalization capability, energy efficiency and low latency advantages seen in nature~\cite{sandamirskaya2022neuromorphic, davies2021advancing, frady2020neuromorphic}.
The fundamental properties of neuromorphic computing that enable its advantageous features encompass highly parallel operations, the integration of processor and memory components, and the utilization of asynchronous event-driven computation with sparse temporal activity~\cite{schuman2022opportunities, sandamirskaya2022neuromorphic, davies2021advancing}.

SNNs represent a set of algorithms within the realm of neuromorphic computing, transferring information through discrete spikes instead of the continuous values used by artificial neural networks~\cite{ghosh2009spiking,gerstner2014neuronal,nunes2022spiking}.
SNN architectures typically comprise interconnected neurons linked via synapses (weights), which transfer information from presynaptic (source) neurons to postsynaptic (target) neurons~\cite{nunes2022spiking}. 
Each neuron has its own internal state, allowing computation at neuron and synapse levels to be parallelized when deployed on neuromorphic hardware, which optimizes data transfer via collocating processing and memory~\cite{frenkel2023bottom, schuman2017survey}.
Deploying SNNs on neuromorphic hardware enables sparse and asynchronous event-driven processing, significantly reducing power consumption and latency~\cite{davies2021advancing,sandamirskaya2022neuromorphic,pei2019towards}.

Various common approaches exist for implementing SNNs, which we mention here to acknowledge their existence. 
One approach is to train an Artificial Neural Network (ANN) using back-propagation and convert the trained model to an SNN for inference.
Strategies for this conversion include activation function approximations~\cite{rueckauer2017conversion}, and optimizations~\cite{bu2021optimal, ding2021optimal}, constrained training to resemble spiking form~\cite{hunsberger2016training, severa2019training}, and optimized spiking neuron models~\cite{stockl2021optimized}. 
Due to limited weight precision, these approaches often do not fully utilize the inherent energy efficiency of SNNs~\cite{schuman2022opportunities}. 
The neuronal dynamics of spiking neurons are non-differentiable which means that back-propagation cannot be directly applied to train SNNs for complex tasks. To address this, a number of works have focused on approximating back-propagation for SNNs~\cite{lee2020enabling, renner2021backpropagation, shen2022backpropagation}. 
However, these methods require offline training with large datasets and perform poorly in continual learning settings due to catastrophic forgetting~\cite{sandamirskaya2022neuromorphic}. 
Lastly, some training paradigms are inspired by the modulation of synaptic strength (weight) based on neuronal activity~\cite{bi1998synaptic}. 
Among these, Spike-Time Dependent Plasticity (STDP) updates weights according to the relative timing of spikes received from presynaptic and postsynaptic neurons~\cite{bi1998synaptic, schuman2022opportunities}. In this study, we employ the unsupervised STDP as our primary training algorithm for SNNs.

We acknowledge the challenges in achieving competitive accuracy with these unsupervised learning approaches. 
To address this, we introduce a novel methodology that combines STDP with a regularization term to detect and remove \textit{hyperactive} neurons, and incorporates enhancement techniques familiar to the conventional machine learning domain, namely modularity, ensembling, and sequence matching. 
This approach aims to enhance SNN accuracy, robustness to significant appearance changes, and scalability in learning a larger number of places.

\subsection{SNNs for Robot Localization}
\label{RW:snns_for_loc}

The capabilities of SNNs have been demonstrated in a wide range of computer vision and robotics tasks. These include pattern recognition~\cite{diehl2015unsupervised}, %
control~\cite{abadia2021cerebellar, vitale2021event, dupeyroux2021neuromorphic, stagsted2020event, ding2022biologically},
manipulation~\cite{tieck2017towards, tieck2018controlling, oikonomou2023hybrid},
object tracking~\cite{lele2021end, luo2022conversion}, 
and scene understanding~\cite{kreiser2020chip, renner2022neuromorphic}. 
Many works have employed SNNs to address tasks related to robot localization, which is the problem under investigation in this work.
These works include 
computational models of place, grid and border cells of the rat hippocampus\cite{galluppi2012live}, 
a navigation controller for mapping unknown environments~\cite{tang2018gridbot}, a pose estimation and map formation method~\cite{kreiser2018pose}, a light-weight system for uni-dimensional SLAM~\cite{tang2019spiking}, and a SLAM model that utilizes representations of continuous spatial maps to produce compressed structures from multiple domains~\cite{dumont2023exploiting}. In previous work~\cite{Hussaini2022}, we presented a SNN specifically for VPR. This network had a limited capacity, recognizing up to just 100 places.

While some of these systems have been deployed on neuromorphic hardware~\cite{galluppi2012live, kreiser2018pose, tang2019spiking, kreiser2018neuromorphic, kreiser2020error}, 
so far, the performance of SNN-based methods for robot localization have only been demonstrated in 
simulated environments~\cite{tang2018gridbot, kreiser2018pose, dumont2023exploiting}, 
constrained indoor environments~\cite{galluppi2012live, tang2019spiking, kreiser2018neuromorphic, kreiser2020error, safa2023fusing}, 
or small-scale outdoor environments~\cite{Hussaini2022}.

In addition to spike-based \emph{processing}, the use of event-based cameras for spike-based \emph{sensing} has shown promising advantages in robotic navigation systems~\cite{milford2015place, weikersdorfer2013simultaneous, vidal2018ultimate}, SLAM systems~\cite{kreiser2018neuromorphic, kreiser2020error, safa2023fusing}, and place recognition~\cite{fischer2022many}, owing to their unique ability to output asynchronous pixel-level brightness changes rather than conventional images, having a high dynamic range and remaining unaffected by motion blur~\cite{gallego2020event}. 
Although our research presently focuses on conventional image frames, the exploration of event-based cameras in the literature highlights the potential of neuromorphic computing in robotics navigation, which is further discussed in~\Cref{discussions_conclusions} as part of our future works.

\subsection{Visual Place Recognition (VPR)}
\label{RW:vpr}

The VPR task is to determine whether a place has been previously visited, even when faced with appearance changes and perceptual aliasing~\cite{Lowry2015, masone2021survey, zhang2021visual, garg2021your, schubert2023visual, tsintotas2022revisiting}.
VPR is most-commonly framed as an image retrieval problem, where feature representations of a given query image are compared to the feature representations of all previously visited places, i.e.~the reference database. The predicted place of the query image is the true position of the reference place that is most similar to the query in feature space. 
The VPR problem can also be posed as a template matching problem, similar to an image classification problem, where associated templates of all reference images are extracted to represent each place either via a single image~\cite{xu2020probabilistic, Arandjelovic2018, cummins2008fab, torii201524} or a sequence of images~\cite{milford2012seqslam, garg2021seqnet, doan2019scalable}.

Learning-based approaches are dominating in VPR. Notably, NetVLAD~\cite{Arandjelovic2018}, a method based on Vector of Locally Aggregated Descriptors (VLAD)~\cite{jegou2010aggregating}, has been influential in producing robust feature representations. Recent advances partitioned training datasets into smaller segments, similar to our approach, and employed an ensemble of classifiers for each segment, facilitating large-scale VPR. For instance, the `Divide and Classify' method \cite{trivigno2023divide} divides the reference dataset into non-overlapping classes. Each segment is processed by an individual classifier, and the collective ensemble is employed for predictions, enabling fast inference for large-scale VPR. On the other hand, Cosplace~\cite{berton2022rethinking} reimagines the training process, viewing it as a classification task and sidestepping the resource-intensive process of mining positive and negative samples inherent in contrastive learning. Notably, both these works \cite{trivigno2023divide, berton2022rethinking} and ours share a common thread: framing VPR as a classification task to further scale place recognition capabilities.

We now review hierarchical approaches to demonstrate the broader landscape of methods aimed at improving recognition accuracy and robustness in VPR, providing a foundation for our exploration of ensembling techniques. Hierarchical techniques have been previously used for coarse-to-fine refinement frameworks via a monolithic neural network to efficiently predict hierarchical features (HF-Net)~\cite{sarlin2019coarse}, probabilistic approaches~\cite{xu2020probabilistic}, bio-inspired approaches~\cite{fan2017biologically}, multi-process fusion~\cite{hausler2020hierarchical},
global-to-local VPR pipeline to guide local feature matching via global descriptors~\cite{keetha2021hierarchical}, 
and hierarchical decomposition of the environment~\cite{garcia2017hierarchical}. In the latter approach, places with similar visual features are grouped together in nodes to reduce search space while maintaining high accuracy~\cite{garcia2017hierarchical}.

\subsection{Non-spiking Biologically Inspired VPR Approaches}
\label{RW:bbio-inspired_vpr}

Biologically inspired VPR methods seek to emulate the navigational capabilities of animals with relatively small brains such as ants, bees, and rodents~\cite{milford2004ratslam, fan2017biologically, neubert2019neurologically, yu2019neuroslam, chancan2020hybrid, bing2023towards} to design energy-efficient and high-performing solutions. 
These non-spiking biologically inspired techniques offer valuable reference points for our spike-based work. 

The place cells and head direction cells in rodent hippocampus inspired RatSLAM~\cite{milford2004ratslam}. RatSLAM was later extended to include grid cells in~\cite{fan2017biologically} and extended to 3D in \mbox{NeuroSLAM}~\cite{yu2019neuroslam}. 
Inspired by the principles of Hierarchical Temporal Memory related to the human neocortex, \cite{neubert2019neurologically} details a minicolumn network to pool spatial information and preserve temporal memory.
\cite{chancan2020hybrid} combines a pattern recognition module, inspired by fruit flies olfactory neural circuit, with a one-dimensional continuous attractor network serving as the temporal filter.
Drawing on head direction cell mechanisms,~\cite{bing2023towards} details a calibration method to correct head direction errors from path integration via visual landmarks.

\subsection{Ensembles of Neural Networks}
\label{RW:ensembling}

One well-known approach to improve the predictive performance of neural networks is to use an ensemble of models~\cite{ghosh2009spiking, ganaie2022ensemble, yang2023survey, dietterich2000ensemble, zhou2012ensemble, sagi2018ensemble}. Ensembles have been shown to generalize well and prevent issues such as over-fitting and instability, which makes these approaches suitable to a wide range of applications in different domains~\cite{ghosh2009spiking, ganaie2022ensemble, yang2023survey, dietterich2000ensemble, zhou2012ensemble, sagi2018ensemble}. 
Challenges in deploying ensembles include requiring sufficient diversity in the output of the individual ensemble members, the trade-off between the computational complexity and the number of ensemble members, and the predictive performance and latency of the ensemble~\cite{ghosh2009spiking, ganaie2022ensemble, yang2023survey, dietterich2000ensemble, zhou2012ensemble, sagi2018ensemble}. 
Although ensembles are typically limited in scalability, our work anticipates leveraging neuromorphic computing, which has the potential to have highly efficient parallelization capability~\cite{davies2021advancing}. Consequently, neuromorphic deployment can alleviate scalability concerns in our ensembling approach due to the small size of each individual network. 

Ensemble techniques have been used for a wide variety of robotic applications including image segmentation~\cite{tan2019evolving}, gaze estimation~\cite{fischer2018rt}, and uncertainty estimation~\cite{lakshminarayanan2017simple}. 
In the field of SNNs, a variety of ensemble methods have been applied to diverse pattern recognition tasks~\cite{shim2016unsupervised, yang2022heterogeneous, fu2021ensemble, elbrecht2020evolving, yin2012reservoir, srinivasan2018spilinc}.
These methods include unsupervised ensembles of spiking expectation maximization networks~\cite{shim2016unsupervised}, and ensembles of evolutionary SNN algorithms~\cite{elbrecht2020evolving} that have been effective in digit recognition. 
Additionally, there are heterogeneous ensembles for few-shot online learning~\cite{yang2022heterogeneous}, and SNN ensembles that use a convolutional structure with unsupervised STDP learning~\cite{fu2021ensemble}, and ensembles of Liquid State Machines~\cite{srinivasan2018spilinc} which have been applied to image~\cite{yang2022heterogeneous, fu2021ensemble, srinivasan2018spilinc} and audio recognition~\cite{srinivasan2018spilinc} tasks. Furthermore, reservoir computing ensembles have been explored for multi-object behavior recognition~\cite{yin2012reservoir}.

This work employs an ensemble of modular SNNs. The diversity within the ensemble members is created via variations in the initialization of the learned weights and the unique set of randomly selected distinct places learned by a module. The work most similar to ours is~\cite{panda2017ensemblesnn} which uses an ensemble of SNNs for digit recognition. 
While~\cite{panda2017ensemblesnn} allocates portions of an input image to different ensemble members for learning, our method processes full images across all modules, whereby each module is trained on a geographically distinct subset of reference places. All module responses are then fused for predicting the corresponding reference place of a given query image.

\subsection{Sequence Matching Techniques for Neural Networks}
\label{RW:seq_matching}

One common approach to improve the robustness of a VPR method, especially against sudden high appearance changes and perceptual aliasing, involves using the temporal information inherent in the database and query sets used in the mobile robot localization task~\cite{masone2021survey, Lowry2015, garg2021your, schubert2023visual, tsintotas2022revisiting}. One such set of algorithms that use the temporal information are sequence matching techniques, which can be classified into similarity-based methods, feature-based methods, and approaches that learn sequential information~\cite{schubert2023visual}.

Similarity-based sequence matching techniques aggregate the similarity scores of a pair of sequences~\cite{schubert2023visual}, which is advantageous as these methods can be used as a filtering process to single-image VPR methods. Similarity-based sequence matching techniques have been developed via local velocity search~\cite{milford2012seqslam}, convolutional operations~\cite{schubert2021fast}, flow network built via a directed acyclic graph~\cite{naseer2014robust}, and Hidden Markov Models~\cite{hansen2014visual}.
However, similarity-based sequence matching techniques do not consider the underlying feature representation method and do not incorporate learning mechanisms~\cite{garg2022seqmatchnet}. Furthermore, elimination of single-image high-confidence false matches cannot be guaranteed without additional contextual information, and their scalability tends to increase linearly with the growth in the size of the reference dataset~\cite{garg2021seqnet}. 

Conversely, feature-based techniques integrate a series of single-image descriptors into a unified descriptor to determine the predicted location of a query image, enabling the sequence descriptor to encompass visual information from both the current place and the preceding places in the sequence~\cite{schubert2023visual, garg2021seqnet, arroyo2015towards, garg2020delta}. 
Learning-based approaches generate a single summary sequential descriptor representing a sequence of images~\cite{garg2022seqmatchnet}. They exploit sequential temporal cues via methods including Transformers, and Convolutional-based architectures~\cite{mereu2022learning}, or Long Short-Term Memory architectures~\cite{facil2019condition}.

In this study, we examine the impact of sequence matching across both conventional and SNN-based VPR techniques. We choose SeqSLAM~\cite{milford2012seqslam} for analysis due to its simplicity and compatibility as a post-processing step for single-image VPR techniques.
We provide analysis on the responsiveness of these techniques to sequence matching. We also introduce an indicator to predict the effectiveness of applying sequence matching, offering new insights into the efficacy of sequence matching across conventional and SNN-based VPR techniques.

\section{Methodology}
\label{methodology}

We first introduce the training regime for a single, compact spiking network that represents a small region of the environment (\Cref{pre}). By combining the predictions of these localized networks at deployment time within a modular scheme (\Cref{Modular}) and introducing global regularization (\Cref{hyp_detection}), we enable large-scale visual place recognition. 
We then present and analyze two enhancements to our modular networks: 1) ensembling, where a single place is represented by multiple networks (\Cref{ensembles_of_SNNs}) and 2) sequence matching, where we make use of multiple query and reference images for place matching (\Cref{sequence_matching}).

\subsection{Preliminaries}
\label{pre}
Our Modular SNN approach is homogeneous, i.e.~each module within the Modular SNN has the same architecture and uses the same hyperparameters. The modules differ in their training data. Each module's training data consists of non-overlapping geographically distinct places of the environment. This specialization makes each module an expert in recognizing a small number of places. 
The SNN architecture introduced in this section, which constitutes a single SNN module in our modular approach, follows~\cite{diehl2015unsupervised,Hussaini2022} and is briefly introduced for completeness in this section. We emphasize that we do not claim a novel SNN architecture.

\begin{figure}[t!]
    \centering
    \includegraphics[width=0.8\columnwidth]{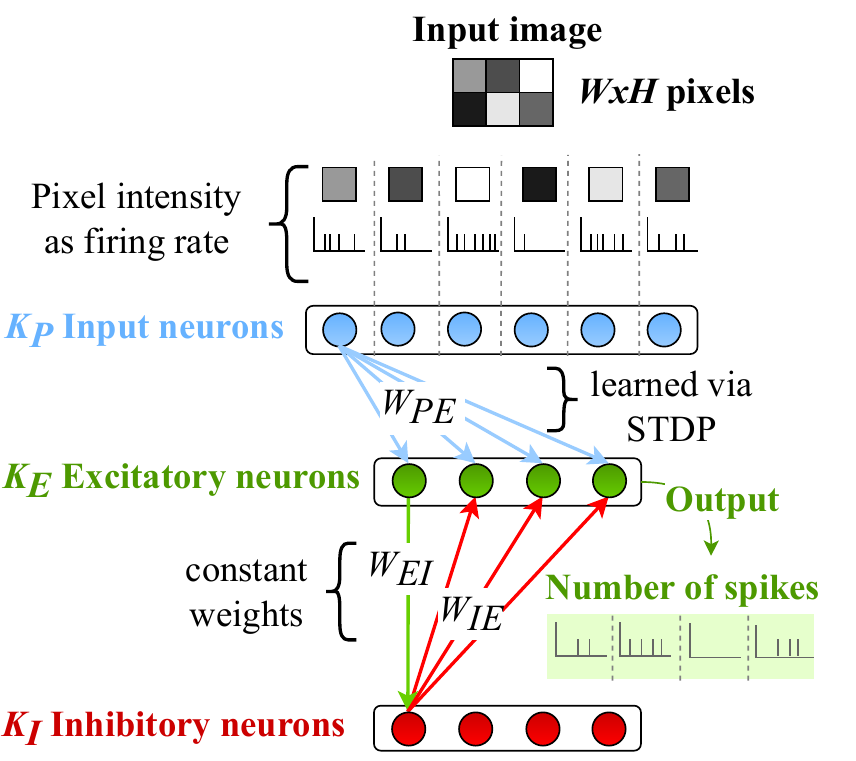}
    \vspace*{-0.25cm}
    \caption{\textbf{SNN Module Architecture:}
    Our Modular SNN is comprised by SNN modules that all have the same three-layer SNN architecture illustrated in this figure. 
    Each module converts an input image to spike trains via rate coding, where the firing rate of input neurons is based on pixel intensities.
    The total number of input neurons is equal to the number of pixels in the input image, denoted as $K_{P}=W \times H$. 
    These input neurons (blue dots) are fully connected to a layer of excitatory neurons (blue arrows connecting to green dots). 
    Each excitatory neuron is connected to a single inhibitory neuron (single green arrow connecting to a red dot), which in turn connects to and inhibits all other excitatory neurons except its paired excitatory neuron (red arrows connecting back to green dots). 
    The synaptic weights from excitatory to inhibitory neurons, $W_{EI}$, and from inhibitory back to excitatory neurons, $W_{IE}$, are fixed constants. 
    The synaptic weights from input neurons to excitatory neurons, $W_{PE}$, are learned via the unsupervised Spike-Time Dependent Plasticity (STDP) mechanism that enables excitatory neurons to respond to different places.
    The number of output spikes from these excitatory neurons is used for place predictions.
    }
    \label{fig:SNN_module}
    \vspace*{-0.4cm}
\end{figure}

\subsubsection{Network Structure}
Each expert module, as illustrated in~\Cref{fig:SNN_module}, consists of a three-layer network architecture: 

\begin{enumerate}[label=\roman*.]
    \item The input layer transforms each input image into Poisson-distributed spike trains via pixel-wise rate coding. The number of input neurons $K_P$ corresponds to the number of pixels in the input image: $K_{P} = W \times H$, where \mbox{$W$ and $H$} correspond to the width and height of the input image respectively. 
    \item The $K_P$ input neurons are fully connected to $K_{E}$ excitatory neurons. Each excitatory neuron learns to represent a particular place, and a high firing rate of an excitatory neuron indicates high similarity between the learned and presented place. Note that multiple excitatory neurons can learn the same place.
    \item Each excitatory neuron connects to exactly one of the $K_{I}$ inhibitory neurons. These inhibitory neurons inhibit all excitatory neurons except the excitatory neuron it receives a connection from. This enables lateral inhibition, resulting in a winner-takes-all system, that generates competition among excitatory neurons for effective learning. 
\end{enumerate}

\subsubsection{Neuronal Dynamics}
The spiking neurons are modeled through their neuronal dynamics using the Leaky-Integrate-and-Fire (LIF) model~\cite{gerstner2014neuronal}. The neuronal dynamics of excitatory neurons' internal voltage is as follows: 
\begin{equation}
\tau_\text{e} \frac{dV}{dt} = (E_{\text{rest},\text{e}} - V) + g_{e} (E_{\text{exc}, \text{e}} - V) + g_{i} (E_{\text{inh},\text{e}} - V), 
\end{equation}
where $\tau_\text{e}$ is neuron time constant, $E_{\text{rest},\text{e}}$ is the resting membrane potential, or the internal voltage when the neuron is not actively receiving spikes, and $E_{\text{exc}, \text{e}}$ and $E_{\text{inh},\text{e}}$ are the equilibrium potentials of the excitatory and inhibitory synapses with synaptic conductance $g_{e}$ and $g_{i}$ respectively. 
The internal voltage of the inhibitory neurons are modeled using the same LIF neuronal dynamics, with neuron time constant $\tau_\text{i}$, resting membrane potential $E_{\text{rest},\text{i}}$, and excitatory and inhibitory equilibrium potentials $E_{\text{exc}, \text{i}}$ and $E_{\text{inh},\text{i}}$.
The equilibrium membrane potentials determine the minimum and maximum internal voltage levels, in this case, depending on their negative or positive sign.

\subsubsection{Network Connections}
The synaptic weights between input neurons and excitatory neurons, $W_{PE}$, are modeled via conductance changes to ensure stable network activity. 
When a synapse receives a presynaptic spike, the synaptic conductance is instantaneously increased by its synaptic weight and the synaptic conductance otherwise decays exponentially, as modeled by: 
\begin{equation}
\tau_{g_{e}} \frac{dg_{e}}{dt} = -g_{e},
\end{equation}
where the time constant of the excitatory postsynaptic neuron is $\tau_{g_{e}}$. The same model is used for inhibitory synaptic conductance $g_{i}$ with the inhibitory postsynaptic potential time constant $\tau_{g_{i}}$.

\subsubsection{Weight Updates}
The synaptic weights between the inhibitory and excitatory neurons are defined with constant synaptic weights, $W_{EI}$ and $W_{IE}$. 
The synaptic weights between input neurons and excitatory neurons, $W_{PE}$, are learned via the biologically inspired unsupervised learning mechanism Spike-Timing-Dependent-Plasticity (STDP). 
The weights are increased if the presynaptic spike occurs before a postsynaptic spike, and decreased otherwise. 
The synaptic weight change $\Delta w_{pe}$ after an input neuron $p$ receives a postsynaptic spike from an excitatory neuron $e$ is defined by: 
\begin{equation}
    \Delta w_{pe} = \eta (\textit{x}_\text{pe,pre} - \textit{x}_\text{pe,tar})(w_\text{max} - w_{pe})^\mu,
\end{equation}
where $\eta$ is the learning rate, $\textit{x}_\text{pe,pre}$ records the number of presynaptic spikes, $\textit{x}_\text{pe,tar}$ is the presynaptic trace target value when a postsynaptic spike arrives, $w_\text{max}$ is the maximum weight, and $\mu$ is a ratio for the dependence of the update on the previous weight.

\subsubsection{Local Regularization of Excitatory Neurons}
To prevent individual excitatory neurons from dominating the response, homeostasis is implemented through an adaptive neuronal threshold. The voltage threshold of the excitatory neurons is increased by a constant $\Theta$ after the neuron fires a spike, otherwise the voltage threshold decreases exponentially. We note that the homeostasis provides regularization only on the \emph{local}, expert-specific scale, not on the \emph{global} modular-level scale.

\subsubsection{Neuronal Assignment}
The network training encourages the network to discern the different patterns (i.e.~places) that were presented during training. As the training is unsupervised, one needs to assign each of the $K_E$ excitatory neurons to one of the $L$ training places ($K_E \gg L$). Following~\cite{diehl2015unsupervised}, we record the number of spikes $S_{e,l}$ of the $e$-th excitatory neuron when presented with an image of the $l$-th place. The highest average response of the neurons to place labels across the local training data is then used for the assignment $A_e$, such that neuron $K_{e}$ is assigned to place $l^*$ if:
\begin{equation}
    A_{e} = l^* = \argmax_{l} S_{e,l}.
\end{equation}

\subsubsection{Place Matching Decisions}
Following~\cite{diehl2015unsupervised}, given a query image $q$, the matched place $\hat{l}$ is the place $l$ which
is the label assigned to the group of neurons with the highest sum of spikes to the query image $\big(\text{i.e.~}A_e = \hat{l}\,\big)$: 

\vspace*{-0.2cm}
\begin{equation}
     \hat{l} = \argmax_{l} \sum_{e[A_e=l]} S_{e,l}^q.
     \label{eq:matching}
\end{equation}
\vspace*{-0.4cm}

\subsection{Modular Scheme} 
\label{Modular}

\subsubsection{Modular Network Structure}
The previous section described how to train individual spiking networks following~\cite{diehl2015unsupervised, Hussaini2022}. In the following sections, we present our novel modular spiking network, which consists of a set of $\mathcal{M}=\{M_1,\dots,M_i,\dots,M_N\}$ expert modules. 
The $i$-th module is tasked to learn the places contained in non-overlapping subsets $R_i \in \mathcal{R}$ of the reference database $\mathcal{R}$, whereby 
\begin{equation}
    \mathcal{R}=\bigcup_{i \in \{1,\dots,N\}} R_i\ \  \text{with}\ R_i \cap R_j = \varnothing\ \ \forall i\neq j.
\end{equation} 
All subsets are of equal size, i.e.~$|R_i|=\kappa$. Therefore, at training time the modules are independent and do not interact with each other. 
This modular approach is aimed at improving the scalability of the individual spiking networks presented in~\Cref{pre}, allowing it to map to a larger number of places than the non-modular approach explored in~\cite{Hussaini2022}.

\subsubsection{Modular Place Matching Decision}
At deployment time, the query image $q$ is provided as input to \emph{all} modules \emph{in parallel}. The predicted place of the Modular SNN, $\hat{l}_{M}$, is obtained by considering the spike outputs of \emph{all} modules, rather than just a single module as in Eq.~(\ref{eq:matching}). We thus refine Eq.~(\ref{eq:matching}) to integrate the contributions of all $N$ modules: 

\vspace*{-0.2cm}
\begin{equation}
     \hat{l}_{M} = \argmax_{l} \sum_{i=1}^{N} \sum_{e[A_e=l]} S_{e,l}^{q, i}.
     \label{eq:matching_mod}
\end{equation}
\vspace*{-0.2cm}

\subsection{Hyperactive Neuron Detection}
\label{hyp_detection}
The basic fusion approach, that considers all spiking neurons of all modules as presented in Eq.~(\ref{eq:matching_mod}), is problematic. As the modules are only ever exposed to their local subset of the training data, there is a lack of global regularization to unseen training data outside of their local subset. In the case of spiking networks, this phenomenon leads to ``hyperactive'' neurons that are spuriously activated when stimulated with images from outside their training data. We decided to detect and remove these hyperactive neurons. This global regularization enhances place recognition capability within the modularity technique, allowing mapping to a larger number of places. 

To detect hyperactive neurons, we do not require access to query data. We feed the entire reference dataset to each SNN module after training, and record the cumulative number of spikes $S_{e,l}^i$ fired by neurons $K_e^i$ of each module $M_i \in \mathcal{M}$ in response to the entire reference dataset $\mathcal{R}$. $S_{e,l}^i$ indicates the number of spikes fired by neuron $K_e$ of module $M_i$ in response to place $l$. 
Neuron $K_e^i$ is considered hyperactive if \begin{equation}
    \label{eq:hyp_thr}
    \sum_l S_{e,l}^i\geq \theta,
\end{equation} 
where $\theta$ is a threshold value that is determined as described in \Cref{ES:Imp_details}. 
The place match is then obtained by the highest response of neurons that are assigned to place $\overline{\hat{l}_{M}}$ after ignoring all hyperactive neurons:

\vspace*{-0.2cm}
\begin{equation}
    \overline{\hat{l}_{M}} = \argmax_l \sum_{i=1}^{N} \sum_{e[A_e=l]} S_{e,l}^{q, i} \mathds{1}_{\sum_l S_{e,l}^i<\theta},
    \label{eq:matching_mod_hyp}
\end{equation}
where the indicator function $\mathds{1}$ filters all hyperactive neurons.

\subsection{Ensemble of Modular SNNs}
\label{ensembles_of_SNNs}

\subsubsection{Ensemble Network Structure} 
In this section, we introduce ensembles of Modular SNNs. The purpose of these ensembles is to improve robustness and generalization ability. The main idea is that each place $l$ is represented by multiple complementary ensemble members. Specifically, the ensemble is represented as a set of 
$\mathcal{E}=\{E_{1},\dots,E_{m},\dots,E_{M}\}$ 
homogeneous ensemble members (i.e.~their network architecture is the same), where each member is an independent Modular SNN. The ensemble members are all trained in parallel on the entire reference database $\mathcal{R}$ (\Cref{fig:frontpage_v2}). 

We generate diversity among the ensemble members through random initialization of learned weights, and random shuffling of the order of input images. This approach aligns with prior work that demonstrated substantial performance improvements in such settings~\cite{lakshminarayanan2017simple}.

\subsubsection{Ensemble Place Matching Decision} 
At deployment time, a query image $q$ is provided as input to all ensemble members in parallel. The predicted place of the Ensemble of Modular SNNs, $\overline{\hat{l}_{E}}$, is determined as the place $l$ which corresponds to the place label that has been assigned to a group of neurons (among all ensemble members) demonstrating the highest cumulative spike activity in response to the input query image $\big(\text{i.e.~}A_e = \overline{\hat{l}_{E}} \,\big)$. Eq.~(\ref{eq:matching_mod_hyp}) is revised as follows to accommodate for the $M$ ensembles members and their corresponding $N$ modules:
\begin{equation}
     \overline{\hat{l}_{E}} = \argmax_l \sum_{m=1}^{M} \sum_{i=1}^{N} \sum_{e[A_e=l]} S_{e,l}^{q, i, m} \mathds{1}_{\sum_l S_{e,l}^i<\theta}.
     \label{eq:matching_ens}
\end{equation}

\subsubsection{Creation of Distance Matrix} 
To compute the single-frame distance matrix for the Ensemble of Modular SNNs, $D_\text{single}$, we compute the cumulative spike activity for all neurons assigned to each place label for each query image $q$.
This contrasts with taking only the maximum that is considered as the prediction in Eq.~\ref{eq:matching_ens}.   
Specifically, given a query image $q$, the $q$-th column of the single-frame similarity matrix $S_\text{single}$ is defined as:

\vspace*{-0.2cm}
\begin{equation}
      S_\text{single}^{(q)} = \sum_{m=1}^{M} \sum_{i=1}^{N} \sum_{e[A_e=l]} S_{e,l}^{q, i, m} \mathds{1}_{\sum_l S_{e,l}^i<\theta}.
     \label{eq:ens_dm}
\end{equation}

We then convert this similarity matrix, $S_\text{single}$, to a distance matrix, $D_\text{single}$, by subtracting the maximum value of the similarity matrix from each element:

\begin{equation}
      D_\text{single}(u, v) = \max(S_\text{single}) - S_\text{single}(u, v).
     \label{eq:convert_distance}
\end{equation}

The same process is applied to our Modular SNN, where all responses from Eq.~\ref{eq:matching_mod_hyp} (rather than just the maximum response used for the prediction) for a given query image $q$ constitute one column of the distance matrix.

\subsection{Sequence Matching}
\label{sequence_matching}

This section briefly introduces the sequence matching technique, SeqSLAM~\cite{milford2012seqslam}, and its convolutional formulation as introduced in SeqMatchNet~\cite{garg2022seqmatchnet}, which we do not claim as our contribution. In \Cref{res:ens_comp_wt_sm}, we will analyze the receptiveness of our Ensemble of Modular SNNs and contrast it to that of conventional techniques. 

Given the single-frame distance matrix $D_\text{single}$ from Eq.~\ref{eq:convert_distance},
we apply the sequence matching operation to the distance value between the reference image at row $v$ and query image at column $u$ to obtain the value of the sequential distance matrix $D_\text{seq}$ at the same corresponding location:

\begin{equation}
\begin{aligned}
    D_\text{seq}(u, v) = {} & \sum_{x \in \{1,\dots,L_{seq}\}} \sum_{y \in \{1,\dots,L_{seq}\}} \\
    & D_{single}(u+x, v+y) I_{L_{seq}}(x,y),
\end{aligned}
\end{equation}
where $I_{L_{seq}}$ is an identity matrix acting as a square filter kernel with dimensions of $L_{seq}\times L_{seq}$ pixels and $L_{seq}$ is the sequence length. 

This sequence matching operation assumes that the reference images and query images are aligned, as convolving the single-frame distance matrix with an identity matrix is a linear temporal alignment process. 
To account for misalignment between the reference and query images, contextual information about the varying speeds between the reference traversal and the query traversal can be used, via Dynamic Time Warping~\cite{xu2022deep} or linear search as done in SeqSLAM~\cite{milford2012seqslam}.

\section{Experimental Setup}
\label{exp_setup}

In this section, we cover our implementation details in~\Cref{ES:Imp_details}, the datasets that we used for evaluation in~\Cref{ES:datasets}, the proof-of-concept robot deployment details in~\Cref{ES:robot_deployment},
and the evaluation metric in~\Cref{ES:evaluation_metrics}. Furthermore, in~\Cref{ES:vpr_methods}, we provide the VPR techniques used for comparison, along with details on the image dimensions in~\Cref{subsec:imagedims}.

\subsection{Implementation Details}
\label{ES:Imp_details}
We implement our SNNs using the Brian2 simulator. We publicly release our code online\footnote{[Online]. Available: \url{https://github.com/QVPR/VPRSNN}}.
Each SNN module contains $K_{E}=400$ excitatory neurons, $K_{I}=400$ inhibitory neurons, and $K_{P}=W \times H$ input neurons, where the input image height and width is $28$ pixels in each dimension. 
We select the network's hyperparameters as follows: For the number of training epochs, we use a fixed value of $30$ epochs for all datasets.
For the threshold value $\theta$ (Eq.~(\ref{eq:hyp_thr})) to detect the hyperactive neurons for each individual Modular SNN evaluated on a dataset, 
we draw the threshold value from a uniformly random distribution with $U(40, 100)$ for small-scale datasets ($<1000$ places) and $U(600, 800)$ for large-scale datasets ($>=1000$ places).
The threshold value of each Modular SNN within an Ensemble of Modular SNNs evaluated on a dataset, is chosen separately based on this uniformly random selection process. 
For our Ensemble of Modular SNNs in the results section,~\Cref{res:I}, we used three or five ensemble members, with the exact number specified in each part.

Following~\cite{diehl2015unsupervised}, we select biologically plausible ranges for all parameters of a single SNN module, except the time constant of the synaptic conductance of the inhibitory neurons, $\tau_{g_{i}}$. We modified $\tau_{g_{i}}$ from $2$ ms, as used by~\cite{diehl2015unsupervised}, to $0.5$ ms. 
The parameter values for a single SNN module are as follows:
For the LIF neuronal dynamics, the voltage threshold of the excitatory neurons is increased by a constant $\Theta = 0.05$ mV after a neuron spikes.
The equilibrium potentials are $E_{\text{exc},\text{e}} = E_{\text{exc},\text{i}} = 0$ mV for both excitatory and inhibitory neurons, $E_{\text{inh},\text{e}} = -100$ mV for excitatory neurons and $E_{\text{inh},\text{i}} = -85$ mV for inhibitory neurons.    
For excitatory neurons, the $\tau_\text{e}$ time constant is $100$ ms, and the resting membrane potential, $E_{\text{rest}, \text{e}}$, is $-65$ mV.  
For inhibitory neurons, the $\tau_\text{i}$ time constant is $10$ ms, and the resting membrane potential, $E_{\text{rest}, \text{i}}$, is $-60$ mV.  
The time constant of the synaptic conductance for the excitatory neurons is $\tau_{g_{e}} = 1$ ms.
For STDP learning, the ratio for the dependence of update on the previous weight is $\mu = 1$, and the maximum weight is $w_\text{max} = 1$.
The learning rate $\eta$ is $1e-4$ if the presynaptic input neuron fires before the postsynaptic excitatory neuron, and $1e-2$ if the reverse occurs.    
The constant synaptic weights between inhibitory and excitatory neurons are $W_{EI} = 10.4$ and $W_{IE} = 17.0$.

We train each SNN module in our Modular SNN with $R_{i}=25$ images, and the total number of SNN modules depends on the size of the reference dataset. 
For instance, given a dataset with $\mathcal{R}=1000$ reference places, we assign each of our SNN modules to learn $R_{i}=25$ distinct places, resulting in a total of $40$ SNN modules. 
For each dataset, we train our SNN-based approach only on the reference set without any pre-training, and use the corresponding query set for testing.
We use QUT's High Performance Computing (HPC) to run each SNN module in a separate CPU job. Most jobs were run on an Intel Xeon Gold 6140 CPU.

\subsection{Datasets}
\label{ES:datasets}

\begin{figure*}[t]
  \centering
  \includegraphics[width=0.78\linewidth]{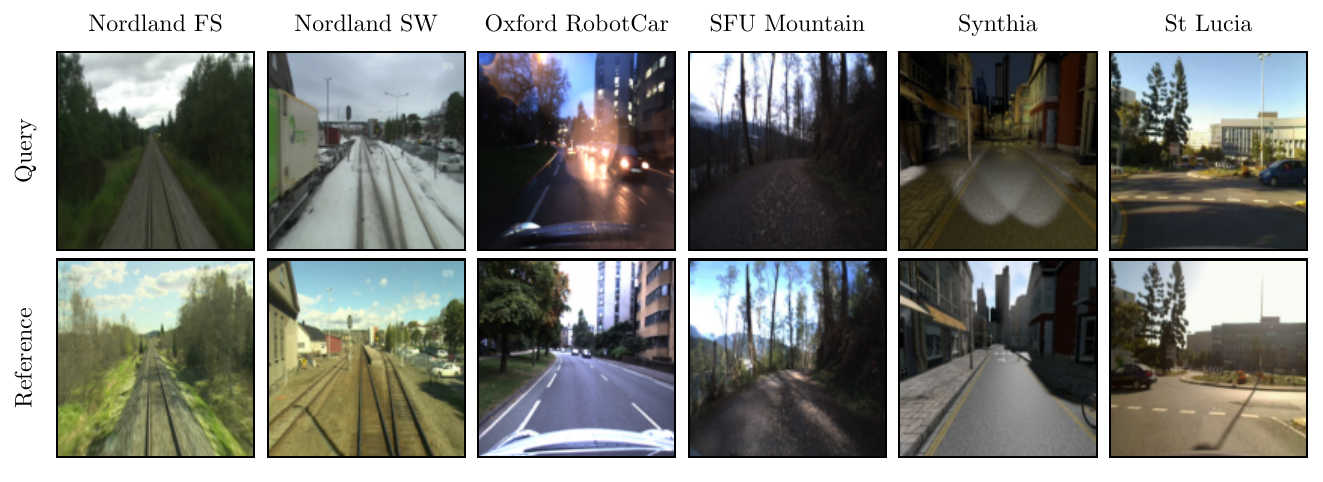}
  \vspace*{-0.25cm} %
  \caption{\textbf{Sample images from the six VPR datasets employed in our research:}
  These datasets encompass a diverse range of environments including urban locales undergoing seasonal transitions, varying illuminations from day to night, high-glare-induced illumination shifts, scenes with occlusions, railway lines, and forested areas.}
  \vspace*{-0.25cm} %
  \label{fig:data_samples}
\end{figure*}

We evaluate our work using several datasets that cover seasonal changes~\cite{sunderhauf2013we,ros2016synthia}, differences in the time of day~\cite{RobotCar,bruce2015sfu,ros2016synthia}, as well as rural~\cite{bruce2015sfu,sunderhauf2013we}, suburban~\cite{milford2008mapping} and urban~\cite{RobotCar,ros2016synthia} environments with additional challenges due to occlusions and glare~\cite{milford2008mapping}. 
We now briefly describe these datasets, and provide sample images in \Cref{fig:data_samples}. 
Our code base\footnote{[Online]. Available: \url{https://github.com/QVPR/VPRSNN}} provides details on how these datasets are used.

\subsubsection{The Nordland dataset~\cite{sunderhauf2013we}} captures a 728 km train path in Norway recorded in spring, summer, fall and winter. As commonly done in the literature~\cite{hausler2021patch, camara2020visual, hausler2019multi}, the data segments where the speed of the train is below 15km/h is removed using the provided GPS data. 
We used the Nordland dataset configured into two different sets: 

\begin{enumerate}[label=\roman*.]
    \item Reference: Fall; query: Summer (also referred to as Nordland FS);
    \item Reference: Spring; query: Winter (also referred to as Nordland SW).
\end{enumerate}

We considered every image in a traverse as a distinct place, obtaining 27575 places for each traverse. 

\subsubsection{The Oxford RobotCar dataset~\cite{RobotCar}} has over 100 traverses captured in Oxford city and it is recorded at varying conditions including different time of the day, and different seasons. 
As done in prior works~\cite{molloy2020intelligent, Hussaini2022}, we selected the front left stereo frames from the Rain (2015-10-29-12-18-17) traverse as the reference dataset and from the Dusk (2014-11-21-16-07-03) traverse as the query dataset.
For each traverse, we sampled approximately one image per meter, resulting in 3800 places. 

\subsubsection{The SFU-Mountain dataset~\cite{bruce2015sfu}} has more than 8 hours of trail driving in Burnaby Mountain, British Columbia Canada, using the Clearpath Husky robot covering sunny, rainy and snowy conditions~\cite{bruce2015sfu}. 
Following prior works~\cite{neubert2021hyperdimensional, lowry2018lightweight}, we used the front right stereo frames from the dry traverse for reference and from the dusk traverse for query. 
We considered each image in the traverse as a different place, and used the author's place sampling configurations~\cite{bruce2015sfu}, using the entire dataset where each traverse contains 375 places.

\subsubsection{The Synthia Night-to-Fall dataset~\cite{ros2016synthia}} is a synthetic dataset that was initially designed for semantic scene understanding in a city-like driving scenario. 
In our approach, following the methodology described in \cite{zaffar2021vpr}, we used the SEQS-04 foggy (reference) and nighttime (query) traverses. Segments in which the vehicle remained stationary were excluded, and we sampled approximately one frame per meter resulting in 250 places for each traverse.

\subsubsection{The St Lucia dataset~\cite{milford2008mapping}} comprises several traverses along a route within the St Lucia suburb of Brisbane. For our experiments, we employed a traverse conducted during the early morning (190809-0845) as the reference traverse and another in the afternoon (180809-1545) as the query traverse. We omitted the segments where the vehicle was at rest and sampled places approximately every 15 meters. We used only the unique places from the reference traverse, obtaining 500 places. For the query traverse, we included sections where places were visited multiple times within the same traverse, obtaining 1037 places.

\subsection{Proof-of-Concept Robot Deployment Setup}
\label{ES:robot_deployment}

For our proof-of-concept CPU-based robot deployment experiment of our Modular SNN, we used AgileX's Scout Mini robot~\cite{agilex_scout_mini} to navigate the QUT Centre for Robotics floor. This area is a shared space with other researchers, where there are moving people, and robots, as well as relocated objects. The robot was tele-operated at a speed of approximately 1 m/s. It collected reference and query traverses at a frequency of 1~Hz at different times of the day, encountering challenges such as occlusions, brightness changes, and slight lateral and frontal viewpoint changes. 
In this experiment, we used a 32GB Intel Core i7 CPU for processing, and Intel’s RealSense D435 for capturing images of the environment.

\subsection{Evaluation Metrics}
\label{ES:evaluation_metrics}

We evaluate the performance of our SNN-based approaches and baseline methods on all datasets using the recall at $N$ (R@$N$) evaluation metric.
This performance metric considers a prediction as a correct match if at least one of the top $N$ predictions is correct~\cite{schubert2023visual, zaffar2021vpr}. 
We deem a query image as correctly paired only if it aligns \emph{exactly} with the correct reference place, employing a ground truth tolerance of zero.  
In the case of sequence matching, the query sequence has to match exactly to the reference sequence.

Let $P_k$ be the set of top $N$ predictions for the $k^{th}$ query and $G_k$ be the set of ground truth matches for the $k^{th}$ query with zero ground truth tolerance. Let $\mathcal{Q}$ be the total number of queries. The recall at $N$ (R@$N$) can be defined as:

\vspace*{-0.2cm}
\begin{equation}
\text{R@}N = \frac{1}{\mathcal{Q}} \sum_{k=1}^{\mathcal{Q}} \mathds{1}\left( P_k \cap G_k \neq \emptyset \right),
\end{equation}
where $\mathds{1}(\cdot)$ is the indicator function, which is $1$ if at least one of the top $N$ predictions for the $k^{th}$ query, $P_k$, is correctly matched to its ground truth $G_k$, and $0$ otherwise.

\subsection{Baseline Methods}
\label{ES:vpr_methods}

We employ several conventional VPR approaches~\cite{milford2012seqslam,torii201524,Arandjelovic2018,revaud2019learning,leyva2023data,berton2022rethinking,ali2023mixvpr} to evaluate the performance of our methods, as well as a comparison to a previous Non-modular SNN~\cite{Hussaini2022} for VPR. These approaches are detailed as follows: 

\subsubsection{Sum-of-Absolute-Differences (SAD)~\cite{milford2012seqslam}} is a simple baseline technique which computes the pixel-wise difference between each query image and all reference images.
While SAD method is simple, this method is not very robust to drastic varying conditions such as changes in lighting, viewpoint, and seasonal changes.

\subsubsection{NetVLAD~\cite{Arandjelovic2018}} aggregates local image descriptors with learnable aggregation weights by employing the VLAD technique~\cite{jegou2010aggregating} to create a fixed-size global descriptor. 
This model has a VGG16~\cite{simonyan2014very} backbone pretrained on ImageNet~\cite{russakovsky2015imagenet}. The NetVLAD layer, which is the pooling layer of the network, is trained on the Google Landmarks~\cite{noh2017large}, Mapillary Street Level Sequences~\cite{warburg2020mapillary}, and Pittsburgh~\cite{torii2013visual} datasets separately. As NetVLAD is trained on urban environments, the model might not generalize well to completely different types of environments such as rural areas. 

\subsubsection{DenseVLAD~\cite{torii201524}} utilizes densely sampled Scale Invariant Feature Transform (SIFT)~\cite{lowe2004distinctive} image descriptors, and aggregates these features using VLAD~\cite{jegou2010aggregating}. While DenseVLAD is very robust to high illumination, some limitations of DenseVLAD include lack of robustness to occlusions, very dark conditions with limited dynamic range, and rural areas with vegetation. 

\subsubsection{AP-GeM~\cite{revaud2019learning}} employs the Generalized-Mean pooling layer (GeM)~\cite{radenovic2018fine} and uses a listwise loss formulation that directly optimizes for the Average Precision (AP) performance metric. 
This model uses a CNN-based backbone pretrained on ImageNet~\cite{russakovsky2015imagenet} to extract feature representations and aggregates it into a compact representation. We used three variations of AP-GeM; a residual networks (ResNet)50 backbone~\cite{he2016deep} trained on Landmarks-clean~\cite{gordo2016deep} dataset, a ResNet101 backbone~\cite{he2016deep} trained on Landmarks-clean~\cite{gordo2016deep} dataset, and a ResNet101 backbone~\cite{he2016deep} trained on Google-Landmarks~\cite{noh2017large} Dataset. Similar to NetVLAD, AP-GeM generalization is reliant on the type of environment used for training. 

\subsubsection{Generalized Contrastive Loss (GCL)~\cite{leyva2023data}} is trained via a GCL using graded similarity labels for image pairs. We trained the last two layers of the network on the Mapillary Street-Level Sequences (MSLS) dataset~\cite{warburg2020mapillary} using a VGG16 backbone~\cite{simonyan2014very} pretrained on ImageNet~\cite{russakovsky2015imagenet} with GeM~\cite{radenovic2018fine} as the pooling layer. 

\subsubsection{CosPlace~\cite{berton2022rethinking}}
uses a classification framework to train the model. 
This approach splits the training dataset into square geographical cells using the ground truth data. During training, the network iterates over CosPlace Groups, which are non-overlapping classes that are grouped together. 
The network uses the Large Margin Cosine Loss (LCML)~\cite{wang2018cosface} with a fully connected layer, that is only present at training, for each group of the dataset. The network consists of a CNN backbone (a VGG-16 backbone), a GeM pooling layer, and a fully connected output layer. At inference, the model outputs compact discriminative feature descriptors. 

\subsubsection{MixVPR~\cite{ali2023mixvpr}} is a global feature aggregation method, that takes the feature maps of a intermediate layers of a CNN-based backbone (a ResNet backbone), that are processed by a cascade of Feature Mixer layers, comprised of multi-layer perceptrons, that provide each element of the feature map with global relationships to all other elements. MixVPR is a state-of-the-art VPR technique across multiple VPR benchmark datasets. 

\subsubsection{Non-modular SNN~\cite{Hussaini2022}} is a single three-layer SNN model that is enlarged to accommodate for learning a significantly larger number of places. The model is trained on the reference dataset, and tested on the query dataset. This comparison is included to show the scalability advantages of ensembling and modularity in terms of both performance and computational time as the number of places to learn increases.

\subsection{Image Dimensions for Different Techniques}
\label{subsec:imagedims}

For our SNN-based methods, we performed the following pre-processing steps: resized each input image to $28 \times 28$ pixels, converted the images to grayscale, applied patch normalization by dividing the images into $7 \times 7$ pixel patches, and normalizing each patch to a range of $-1$ to $1$ using its mean and standard deviation. Finally, we scaled the pixel values to be between $0$ and $255$.
For GCL and NetVLAD, the images were resized to $640 \times 480$ pixels, 
while for AP-GeM, DenseVLAD, CosPlace, and MixVPR %
the native image resolutions were used (Nordland: $640 \times 360$ pixels, Oxford RobotCar: $1280 \times 960$ pixels, SFU-Mountain: $752 \times 480$ pixels, Synthia Night To Fall: $300 \times 200$ pixels, St Lucia: $640 \times 480$ pixels). 
For SAD, the input images were resized to $28 \times 28$ pixels and patch-normalized with patch sizes of $7 \times 7$ pixels, matching the low-dimensional input image sizes that we used in our work.

\section{Results}
\label{res:I}

\Cref{res:overview_ens_seq} analyzes how each component of our methodology affects the overall performance of the system. Subsequently, \Cref{res:ens_comp_no_sm} compares the performance of our Ensemble of Modular SNNs \emph{without} sequence matching against conventional VPR techniques.  \Cref{res:ens_comp_wt_sm} extends this comparison to include sequence matching. 
\Cref{res:Ind_seq} provides detailed analyses and introduces an indicator to assess the responsiveness of VPR techniques to sequence matching. 
We also provide an evaluation of the ensembling effect on our Modular SNN in \Cref{res:EnsModSNN}.
This is followed by an ablation study on the effect of ensemble member randomization in our Ensemble of Modular SNNs in \Cref{res:abl_ModSNN_rand}. Additionally, \Cref{res:compute_scalability} provides an analysis of the computational efficiency and scalability aspects of our approach.
Finally, \Cref{res:robot_deployment} concludes this section with a proof-of-concept robot deployment in a small indoor environment.

\begin{figure}[t]
  \centering
  \includegraphics[width=0.8\linewidth]{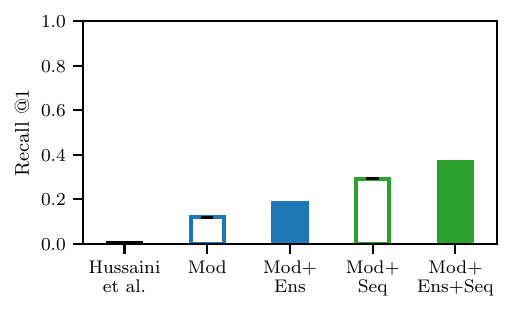}
  \vspace*{-0.2cm}
  \caption{
  \textbf{Component-wise ablation study:}
  Introducing modularity (Mod) where multiple SNNs represent small subsets of the reference dataset enables large-scale place recognition, significantly outperforming the Non-modular baseline SNN by Hussaini et al.~\cite{Hussaini2022}. Both ensembling (five ensemble members; Mod+Ens) and sequence matching (sequence length four; Mod+Seq) individually enhance the R@1 of the Modular SNN, by 6.3\% and 17.2\% respectively. Their combined application (Mod+Ens+Seq) further elevates the performance, surpassing the benefits of the individual techniques and resulting in a 24.9\% R@1 improvement overall. Error bars indicate performance variations among the five ensemble members (standard deviation). 
  The experiment was conducted on the Nordland dataset (Reference: Spring, Fall; query: Winter).
  }
  \label{fig:ens_seq_abl}
\end{figure}

\subsection{Component-Wise Contributions to SNN Performance}
\label{res:overview_ens_seq}

This section overviews the performance contributions of each component of our approach, namely modularity (i.e.~expert modules that learn small subsets of the reference dataset), ensembling (i.e.~representing each place by multiple modules), and sequence matching (i.e.~using multiple reference and query images for place matching).
\Cref{fig:ens_seq_abl} demonstrates that the performance of our Modular SNN is significantly increased when each of these techniques, ensembling (R@1 increase of 6.3\% with five ensemble members) and sequence matching (R@1 increase of 17.2\% for a sequence length of four), is applied separately. Moreover, the combination of ensembling and sequence matching techniques further improves the R@1 of the Modular SNN  (R@1 increase of 24.9\%). As the ensembling and sequence matching techniques are commutative, the order of application of these two techniques produces identical outcomes.   
The Non-modular SNN by Hussaini et al.~\cite{Hussaini2022} performs poorly, even when ensembling and sequence matching are applied (R@1 is less than 0.2\%).

\begin{table*}[t]
\caption{R@1 performance comparisons of our Modular SNN and Ensembles of Modular SNNs to conventional VPR techniques with a sequence matcher at different sequence lengths (SL): SL1 (without a sequence matcher), SL2, SL4, and SL10. 
The main key takeaways include: Our Ensemble of Modular SNNs 
1) shows competitive performance, 
comparable to multiple VPR methods across various datasets, 
2) obtains the highest R@1 improvement with a sequence matcher, compared to VPR techniques with similar-performing baselines except on Oxford RobotCar, and 
3) consistently outperforms the mean R@1 of its individual members across all datasets.} 
\label{tab:RAt1_comparisons}
\resizebox{\textwidth}{!}{%
\setlength{\tabcolsep}{2.5pt}
\renewcommand{\arraystretch}{1}
\begin{tabular}{l|cccc|cccc|cccc|cccc|cccc|cccc|cccc}
\textbf{Datasets} & \multicolumn{4}{c|}{\textbf{Nordland FS}} & \multicolumn{4}{c|}{\textbf{Nordland SW}} & \multicolumn{4}{c|}{\textbf{Oxford RobotCar RD}} & \multicolumn{4}{c|}{\textbf{SFU Mountain}} & \multicolumn{4}{c|}{\textbf{Synthia Night to Fall}} & \multicolumn{4}{c|}{\textbf{St Lucia}} & \multicolumn{4}{c}{\textbf{Mean R@1}}\\ 
\hline
\textbf{Method}  & SL1  & SL2  & SL4  & SL10  & SL1  & SL2  & SL4  & SL10  & SL1  & SL2  & SL4  & SL10 & SL1  & SL2  & SL4  & SL10 & SL1  & SL2  & SL4  & SL10 & SL1  & SL2  & SL4  & SL10 & SL1  & SL2  & SL4  & SL10\\
\hline

NetVLAD (Landmarks)       & 0.57 & 0.65 & 0.73 & 0.80 & 0.12 & 0.16 & 0.21 & 0.30 & 0.08 & 0.10 & 0.14 & 0.22 & 0.40 & 0.48 & 0.62 & 0.87 & 0.79 & 0.84 & 0.87 & 0.87 & 0.26 & 0.45 & 0.60 & 0.79 & 0.37 & 0.45 & 0.53 & 0.64\\
NetVLAD (Mapillary)       & 0.50 & 0.60 & 0.70 & 0.78 & 0.10 & 0.14 & 0.20 & 0.31 & 0.08 & 0.10 & 0.14 & 0.24 & 0.32 & 0.42 & 0.53 & 0.76 & 0.67 & 0.76 & 0.87 & 0.88 & 0.30 & 0.42 & 0.59 & 0.79 & 0.33 & 0.41 & 0.51 & 0.63\\
NetVLAD (Pittsburgh)      & 0.43 & 0.54 & 0.65 & 0.76 & 0.08 & 0.11 & 0.14 & 0.21 & 0.12 & 0.15 & 0.21 & 0.33 & 0.39 & 0.56 & 0.71 & 0.90 & 0.81 & 0.86 & 0.89 & 0.87 & 0.26 & 0.43 & 0.57 & 0.78 & 0.35 & 0.44 & 0.53 & 0.64\\
AP-GeM (ResNet-50)        & 0.47 & 0.56 & 0.66 & 0.75 & 0.07 & 0.10 & 0.13 & 0.19 & 0.11 & 0.13 & 0.18 & 0.29 & 0.40 & 0.53 & 0.75 & 0.92 & 0.59 & 0.76 & 0.83 & 0.87 & 0.35 & 0.57 & 0.78 & 0.95 & 0.33 & 0.44 & 0.55 & 0.66\\
AP-GeM (ResNet-101)       & 0.48 & 0.58 & 0.67 & 0.76 & 0.06 & 0.08 & 0.09 & 0.13 & 0.12 & 0.15 & 0.19 & 0.31 & 0.43 & 0.58 & 0.75 & 0.95 & 0.62 & 0.75 & 0.84 & 0.85 & 0.34 & 0.54 & 0.72 & 0.90 & 0.34 & 0.45 & 0.54 & 0.65\\
AP-GeM (ResNet-101, LM18) & 0.67 & 0.74 & 0.78 & 0.82 & 0.20 & 0.26 & 0.34 & 0.47 & 0.12 & 0.15 & 0.20 & 0.29 & 0.61 & 0.76 & 0.88 & 0.96 & 0.73 & 0.84 & 0.87 & 0.87 & 0.55 & 0.75 & 0.91 & 0.98 & 0.48 & 0.58 & 0.67 & 0.73\\
DenseVLAD                 & 0.65 & 0.73 & 0.77 & 0.81 & 0.16 & 0.26 & 0.37 & 0.54 & 0.19 & 0.24 & 0.32 & 0.47 & 0.67 & 0.78 & 0.86 & 0.96 & 0.77 & 0.80 & 0.87 & 0.87 & 0.61 & 0.80 & 0.92 & 0.99 & 0.51 & 0.60 & 0.68 & 0.77\\
SAD                       & 0.52 & 0.61 & 0.69 & 0.78 & 0.15 & 0.22 & 0.30 & 0.44 & 0.19 & 0.25 & 0.29 & 0.37 & 0.41 & 0.51 & 0.63 & 0.80 & 0.48 & 0.61 & 0.68 & 0.78 & 0.24 & 0.32 & 0.41 & 0.40 & 0.33 & 0.42 & 0.50 & 0.60\\
GCL                       & 0.51 & 0.62 & 0.71 & 0.78 & 0.14 & 0.21 & 0.30 & 0.50 & 0.05 & 0.06 & 0.07 & 0.12 & 0.45 & 0.57 & 0.69 & 0.86 & 0.52 & 0.68 & 0.83 & 0.88 & 0.29 & 0.47 & 0.67 & 0.84 & 0.33 & 0.43 & 0.55 & 0.66\\
MixVPR                    & 0.78 & 0.81 & 0.82 & 0.83 & 0.73 & 0.84 & 0.91 & 0.96 & 0.23 & 0.28 & 0.36 & 0.51 & 0.86 & 0.94 & 0.99 & 1.00 & 0.84 & 0.87 & 0.89 & 0.88 & 0.99 & 1.00 & 1.00 & 1.00 & 0.65 & 0.72 & 0.78 & 0.81\\
CosPlace                  & 0.73 & 0.77 & 0.79 & 0.82 & 0.60 & 0.72 & 0.84 & 0.93 & 0.12 & 0.14 & 0.18 & 0.25 & 0.74 & 0.87 & 0.97 & 1.00 & 0.82 & 0.85 & 0.87 & 0.88 & 0.90 & 0.97 & 1.00 & 1.00 & 0.74 & 0.79 & 0.83 & 0.86\\
Modular SNN (Ours)        & 0.42 & 0.55 & 0.67 & 0.78 & 0.12 & 0.19 & 0.29 & 0.47 & 0.15 & 0.21 & 0.28 & 0.40 & 0.36 & 0.56 & 0.78 & 0.95 & 0.35 & 0.49 & 0.64 & 0.82 & 0.24 & 0.38 & 0.57 & 0.84 & 0.27 & 0.40 & 0.54 & 0.71\\

\hline

Ens of 3 AP-GeM              & 0.64 & 0.71 & 0.77 & 0.81 & 0.14 & 0.18 & 0.23 & 0.31 & 0.16 & 0.18 & 0.24 & 0.35 & 0.61 & 0.72 & 0.86 & 0.97 & 0.73 & 0.81 & 0.87 & 0.87 & 0.55 & 0.69 & 0.87 & 0.97 & 0.47 & 0.55 & 0.64 & 0.71\\
Ens of 3 GCL                 & 0.52 & 0.62 & 0.71 & 0.78 & 0.15 & 0.23 & 0.33 & 0.54 & 0.05 & 0.05 & 0.07 & 0.11 & 0.46 & 0.57 & 0.70 & 0.86 & 0.54 & 0.68 & 0.82 & 0.88 & 0.29 & 0.47 & 0.68 & 0.83 & 0.33 & 0.44 & 0.55 & 0.66\\
Ens of 3 Modular SNNs (Ours) & 0.50 & 0.62 & 0.72 & 0.80 & 0.17 & 0.25 & 0.35 & 0.54 & 0.20 & 0.25 & 0.33 & 0.44 & 0.43 & 0.64 & 0.83 & 0.98 & 0.39 & 0.57 & 0.72 & 0.85 & 0.27 & 0.44 & 0.65 & 0.89 & 0.33 & 0.46 & 0.60 & 0.75\\

\hline

Ens of 5 GCL                 & 0.52 & 0.62 & 0.70 & 0.78 & 0.15 & 0.24 & 0.35 & 0.55 & 0.05 & 0.05 & 0.07 & 0.11 & 0.47 & 0.56 & 0.70 & 0.86 & 0.54 & 0.67 & 0.82 & 0.88 & 0.30 & 0.47 & 0.67 & 0.83 & 0.34 & 0.44 & 0.55 & 0.67\\
Ens of 5 Modular SNNs (Ours) & 0.52 & 0.63 & 0.72 & 0.80 & 0.18 & 0.26 & 0.37 & 0.55 & 0.21 & 0.27 & 0.35 & 0.44 & 0.46 & 0.67 & 0.85 & 0.98 & 0.44 & 0.59 & 0.72 & 0.85 & 0.32 & 0.49 & 0.69 & 0.92 & 0.35 & 0.49 & 0.62 & 0.76\\

\end{tabular}
}
\vspace{-0.4cm}
\end{table*}

\subsection{Comparison of Ensemble of Modular SNNs \textbf{without} a Sequence Matcher to Conventional VPR Techniques}
\label{res:ens_comp_no_sm}

This section compares our Ensemble of Modular SNNs without a sequence matcher to the conventional VPR techniques outlined in~\Cref{ES:vpr_methods} on the datasets detailed in~\Cref{ES:datasets}. 
As emphasized in~\cite{schubert2021makes}, the efficacy of different visual place recognition methods fluctuates across different environments. The aim of our work is to demonstrate competitive but not necessarily state-of-the-art performance of our approach for visual place recognition.

Table~\ref{tab:RAt1_comparisons} shows that 
our Ensemble of Modular SNNs consistently delivers competitive results, that are in close proximity to some of the leading VPR methods across various datasets. 
On average across all datasets, the top-performing methods include CosPlace, MixVPR, DenseVLAD, and AP-GeM (ResNet-101, LM18).
It is worth noting that conventional VPR techniques have inherent advantages. They operate on much larger image dimensions as they falter with smaller image sizes which are used in our approach, and oftentimes benefit from extensive pretraining on large VPR datasets.

\begin{figure*}[htbp]
  \centering
  \includegraphics[width=\linewidth]{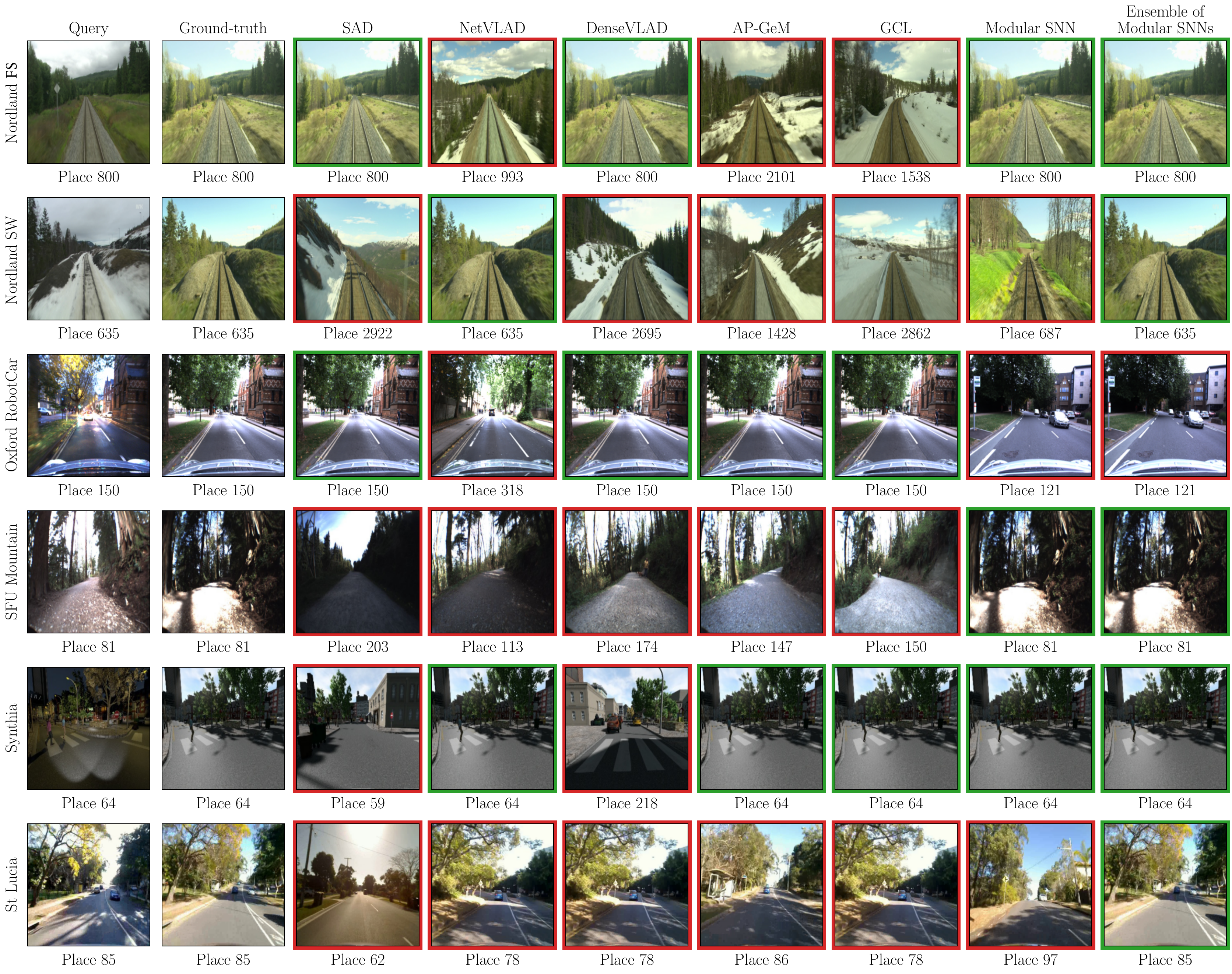}
  \caption{\textbf{Qualitative results:} 
  The plot showcases the performance of our Standalone Modular SNN, Ensemble of Modular SNNs (with five ensemble members), and various VPR methods across diverse datasets. 
  To enhance clarity, we have included just one of the three NetVLAD instances, specifically NetVLAD (Landmarks), and one of the three AP-GeM instances, namely AP-GeM (ResNet101, LM18).
  It details three instances of correct predictions by both Modular SNN and its Ensemble variant (in rows one, four, and five), two cases where the Ensemble yields correct matches despite the incorrect prediction of the Modular SNN as an ensemble member (in rows two and six), and a situation where both Modular and Ensemble of Modular SNNs fail to correctly match the query image to its corresponding reference image (in row three).
  }
  \label{fig:qualitative_res_correct}
  \vspace*{-0.4cm}
\end{figure*}

\Cref{fig:qualitative_res_correct} presents qualitative performance of  our Modular SNN, Ensemble of Modular SNNs, and relevant VPR techniques used for comparison across all datasets, for both correct and incorrect prediction of query image instances.

\begin{figure*}[t]
  \centering

  \subfloat{\includegraphics[width=\linewidth]{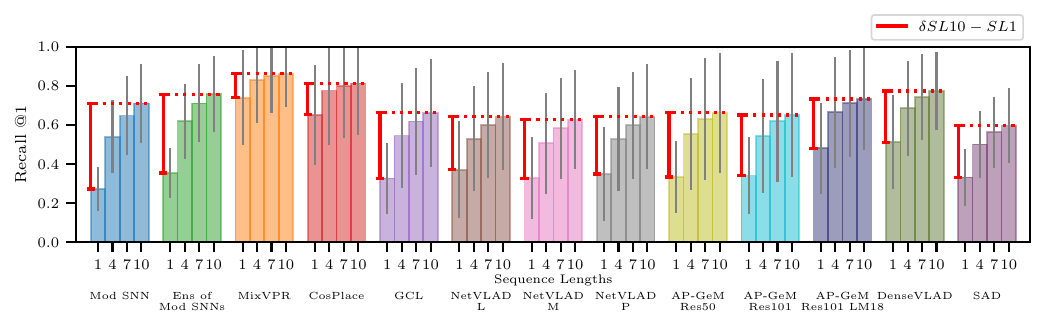}}

  \vspace*{-0.4cm}
  \caption{
  \textbf{R@1 performance improvements with sequence matching:}
  The plot shows the mean R@1 performance of each method across all datasets when employing sequence matching using four, seven and ten frames, compared to the single-frame approach (SL1).
  The gray lines represent the standard deviation of the R@1 model performance across all datasets.
  Red bars demonstrate the mean R@1 performance improvement of a method without a sequence matcher to the performance of the method with a sequence matcher of sequence length ten. 
  Our Modular and Ensemble of Modular SNNs obtain the highest R@1 improvement with a sequence matcher (from without a sequence matcher to a sequence matcher of sequence length ten).  
  The mean R@1 performance of both our Ensemble of Modular SNNs (with five ensemble members), and Modular SNN without a sequence matcher (SL1) is competitive with multiple VPR techniques, and incorporating a sequence matcher with sequence lengths of four, seven, and ten enables our SNN-based approaches to obtain the highest R@1 improvement compared to similar-performing VPR baselines. 
  Notably, the R@1 performance of our SNN-based approaches with a sequence matcher of sequence length ten frames slightly surpasses that of AP-GeM (ResNet101, LM18), and is in close approximately to that of DenseVLAD, both of which have higher-performing baselines (without a sequence matcher).
  }
  \vspace*{-0.4cm}

  \label{fig:bar_plot}
\end{figure*}

\subsection{Comparison of Ensemble of Modular SNNs \textbf{with} a Sequence Matcher to Conventional VPR Techniques}
\label{res:ens_comp_wt_sm}

This section extends the comparisons, this time incorporating a sequence matcher. 
Table~\ref{tab:RAt1_comparisons} shows the performance of our Ensemble of Modular SNNs as well as our standalone Modular SNN when integrated with a sequence matcher with a sequence length of two, four, and ten, \emph{separately}, compared to conventional VPR techniques and/or their ensemble forms with a sequence matcher of same sequence lengths across all datasets. 

Compared to VPR techniques with roughly similar baseline performance, our Ensemble of Modular SNNs (with five ensemble members) obtains the overall highest improvement with a sequence matcher averaged across all datasets: the mean R@1 of our Ensemble of Modular SNNs with a sequence matcher across all datasets is among the top five models, despite the lower baseline performance without a sequence matcher. %
Specifically, our Ensemble of Modular SNNs with a four-frame sequence matcher achieved a 39.1\% R@1 performance gain on the SFU Mountain dataset with a baseline performance of 46.1\%, surpassing the 22.9\% increase of the comparable VPR technique, GCL, with a similar baseline performance of 47.2\%. 

On the Nordland SW dataset, before applying the sequence matcher, the R@1 of our model was 18.3\%, comparable to the baseline performance of AP-GeM (ResNet101, LM18), which stood at 19.6\%. Upon integrating the sequence matcher with four frames, the performance of our model increased to 36.9\%, an increase that slightly surpasses the post-sequence matching performance gains observed in AP-GeM (ResNet101, LM18), which reported an increase to 34.1\%. 

\Cref{fig:bar_plot} illustrates the R@1 of all techniques with a sequence matcher of sequence lengths four, seven, and ten averaged across all datasets. 
With the integration of a sequence matcher of sequence length ten, our Ensemble of Modular SNNs achieves higher R@1 improvements over all other VPR methods. We note that methods such as MixVPR and CosPlace, which have an already significantly higher baselines, compared to all other evaluated techniques, do not significantly benefit from being paired with a sequence matcher.

It is noteworthy that at longer sequence lengths, the R@1 performance of all techniques converge towards perfect R@1 values, which makes it challenging to distinguish between the performances of these techniques with a sequence matcher. 
The variations in absolute performance gain of different techniques are more pronounced in shorter sequence lengths, offering clearer insights into the adaptability. 
Sequence matchers with shorter lengths are apt for indoor settings or high-speed contexts where successive visual scenes change swiftly, and reducing relocalization delays. %

\begin{figure}[t]
  \centering
  \hfill\subfloat{\includegraphics[width=0.95\linewidth,trim={1mm 1mm 1mm 1mm},clip]{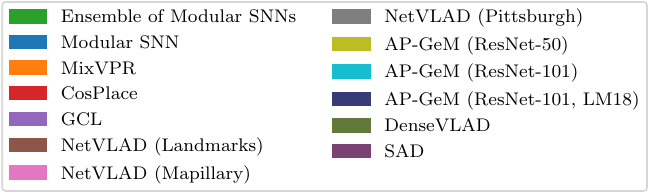}}
  
  \vspace{-0.2cm} %
  
  \hfill\subfloat{\includegraphics[width=\linewidth,trim={1mm 1mm 1mm 1mm},clip]{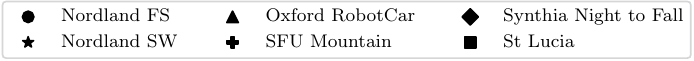}}

  \vspace{-0.4cm} %
  
  \subfloat{\includegraphics{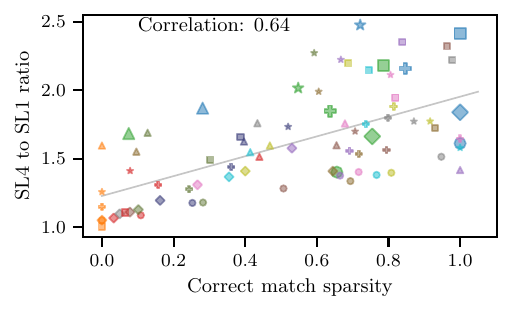}}

  \vspace*{-0.4cm}

  \caption{
  \textbf{Indication of sequence matching responsiveness:}
  The figure shows the correct match sparsity against the R@1 performance ratio of methods with a sequence matcher of sequence lengths four to one. 
  Correct match sparsity is defined as the mean distance to the next correct match for all predictions of query images, projected to log space and min-max normalized for each dataset to allow for dataset agnostic evaluation. 
  Higher sparsity indicates wider gaps between correct matches, suggesting potential under-performance. The gray line represents the best fit line for all data points. 
  Key observations are: 
  1) Our Modular SNN and Ensemble of Modular SNNs (with five ensemble members) generally show a higher SL4 to SL1 ratio compared to nearby data points, regardless of the dataset the methods were evaluated on, indicating strong adaptability to sequence matching. 
  Exceptions include our Ensemble's performance on Nordland FS, Synthia Night to Fall, and our Modular SNN evaluated on Nordland FS, which are comparable to other methods.
  2) Across all data points, methods with high correct match sparsity show greater responsiveness to sequence matching, regardless of the method observed. 
  }
  \vspace*{-0.4cm}

  \label{fig:seq_match_ratio}
\end{figure}

\begin{figure*}[t]
  \centering
  \includegraphics[width=\textwidth]{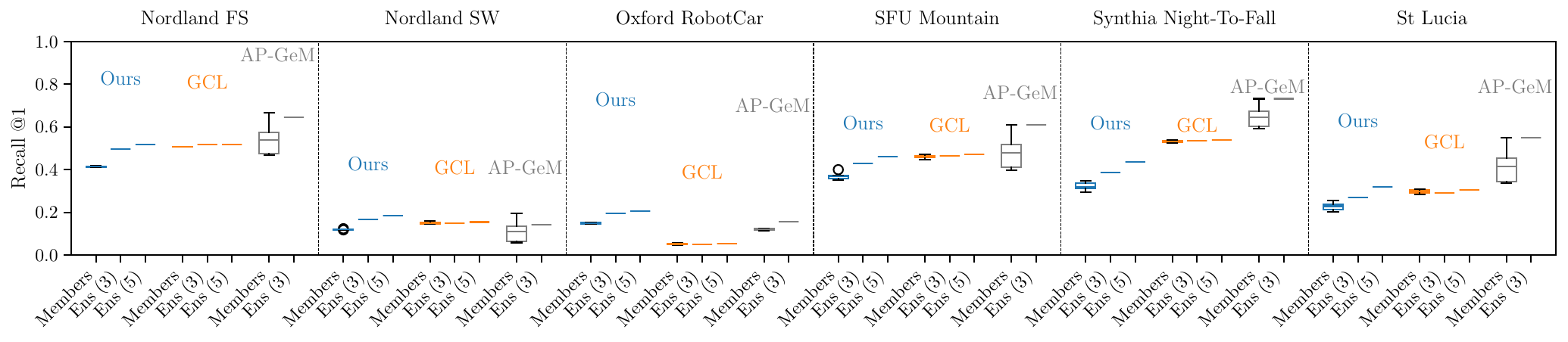}

  \caption{\textbf{The effect of ensembling:}
  The plot shows the R@1 performance of Ensembles of Modular SNNs (both with three, and five ensemble members; blue), GCL ensembles (both with three, and five ensemble members; orange), and an AP-GeM ensemble (with three ensemble members; gray) on all six evaluated datasets. 
  Our Ensemble of Modular SNNs show superior R@1 performance over its individual members, while GCL ensembles exhibit minimal gains. 
  The AP-GeM ensemble members have a varied R@1 performance spectrum, with the AP-GeM ensemble performance matching or falling short of its best-performing member across all datasets. 
  For detailed metrics across all datasets, refer to Table \ref{tab:RAt1_comparisons}.
  }
  \label{fig:ens_Rat1}
  \vspace*{-0.4cm}
\end{figure*}

\subsection{Indicator for Sequence Matching Responsiveness}
\label{res:Ind_seq}

This section examines whether the responsiveness of VPR techniques to sequence matching can be predicted, and offers insights that corroborate their respective behaviors. Specifically, we investigate how the sparsity of correct matches influences the R@1 performance when sequence matching is employed. Increased sparsity in correct matches denotes larger gaps between next correct matches across all predictions of query images due to limitations in the performance of a method.

We define the correct match sparsity as the mean distance to the next correct match for all predictions of query images, projected to log space and min-max normalized per dataset, which enables dataset-agnostic evaluation. 
Let $D$ be the distance matrix of size $ L_{R} \times L_{Q}$, where $L_{R}$ is the number of reference places, and $L_{Q}$ is the number of query places. Further, let $GT$ be a binary matrix of the same size, where a $1$ indicates true matches, and $0$ indicates false matches. The indices of correct matches are identified as follows: 

\vspace*{-0.2cm}
\begin{equation}
\{ q\ |\ q =\argmin_i~D(r, i) \wedge GT(r,i) = 1,\ r = \{1, \dots, L_R\} \}.
\end{equation}
\vspace*{-0.2cm}

The difference sequence \( \Delta q \) can then be defined as:
\vspace*{-0.2cm}
\begin{equation}
\Delta q = \{ \Delta q_1, \Delta q_2, \Delta q_3, \ldots, \Delta q_{n-1} \},
\end{equation}
where $\Delta q_i = q_{i+1} - q_i \quad \text{for} \quad i = \{1, 2, \ldots, n-1\}$.
The mean distance $\tilde{d}$ to the next correct match for all correct matches in the distance matrix is calculated as:

\vspace*{-0.2cm}
\begin{equation}
    \tilde{d} = \frac{1}{|\Delta q|} \sum_{i} \Delta q_i.
\end{equation}
\vspace*{-0.2cm}

\Cref{fig:seq_match_ratio} presents the relationship between the correct match sparsity $\tilde{d}$ and the R@1 performance ratio for a sequence matcher with a length of four against a sequence matcher with a length of one. For better visualization, this figure shows $\log(\tilde{d})$ with a min-max-normalization across different datasets, such that the sparsity ranges between $0$ and $1$.

Central to our discussion, the data points representing our Modular SNN evaluated on different datasets, distinguished by their larger blue points, 
show a higher SL4 to SL1 ratio compared to data points with a similar correct match sparsity, 
emphasizing the robustness and adaptability of our Standalone Modular SNN to sequence matching. An exception is the evaluation on Nordland FS which showed similar performance to nearby data points. 
Similarly, our Ensemble of Modular SNNs, represented by their larger green points, has a high responsiveness to sequence matching, 
with most data points being higher than methods with a similar correct match sparsity. However, on Nordland FS and Synthia Night to Fall, our Ensemble of Modular SNNs has similar performance to nearby methods.

As correct match sparsity correlates to method performance, it means that methods with lower baseline performance generally benefit more from sequence matching compared to methods with higher baseline performance.
Methods with higher-performing baselines include MixVPR and CosPlace, whose data points mainly occupy the lower left of the plot. 
Due to min-max normalization across datasets, low correct match sparsity indicates the highest-performing methods, while high sparsity suggests lower performance within each dataset. 

\subsection{Ensembling: How Much Does It Help?}
\label{res:EnsModSNN}

This section evaluates the effect of ensembling on our Modular SNN, and provides comparisons to GCL and AP-GeM ensembles. The ensemble members in the case of our Modular SNN are homogeneous as they share the same network architecture and training data; their differences lie in the random initialization values of weights and the random sequence of input images, as elaborated in \Cref{ensembles_of_SNNs}. The GCL ensembles are also homogeneous with consistent network architecture and training datasets and only differing in random initialization of weights and order of input images. The AP-GeM ensembles are heterogeneous, showing diverse Convolutional Neural Network (CNN) backbone and/or training datasets. The three architectures are a ResNet50 and a ResNet101, both trained on the Landmarks-clean dataset, and a ResNet101 trained on the Google-Landmarks Dataset, as described in~\Cref{ES:vpr_methods}.

We created the GCL and AP-GeM ensembles by averaging the feature representations of each reference and query set across all ensemble members, and then computed the distance matrix based on these averaged representations\footnote{Additionally, we explored creating these ensembles through element-wise summation of the distance matrices of all ensemble members. However, we selected the averaged feature representation method because it performed better than the method of combining distance matrices.}.

\Cref{fig:ens_Rat1} presents the R@1 performance improvement of our Ensemble of Modular SNNs (both with three, and five ensemble members), ensemble of GCL (both with three, and five ensemble members), and an ensemble of AP-GeM models (with three ensemble members) relative to the mean R@1 of their respective ensemble members across all datasets. 
Individual Modular SNNs perform relatively consistent with little variation in R@1 performance. The Ensemble of Modular SNNs (both with three, and five ensemble members) consistently outperforms the average R@1 of its individual members.
Across all datasets, the ensembles with three members achieve an average R@1 of 32.5\%, compared to their individual members' mean R@1 of 26.6\%. Similarly, ensembles with five members reach an average R@1 of 35.4\%, exceeding their members' mean R@1 of 26.7\%.

While the homogeneous GCL ensemble members achieve consistent R@1 performance, which is similar to the consistency in performance of our Modular SNN ensemble members, the GCL ensemble (both with three, and five ensemble members) shows little to no improvement in R@1 over its individual member average (R@1 improvements of less than 2\%) in both three and five ensemble member instances. 
It is likely that the different GCL ensemble members all converge to the same local minima because of the loss function used in this approach.

The ensemble members of the heterogeneous AP-GeM models exhibit a wide range of R@1 performance. The ensemble performance is equal to and/or inferior to the R@1 of the best-performing ensemble member across all six datasets (see~\Cref{tab:RAt1_comparisons}).
Across all datasets, the mean R@1 for the ensemble is 47.3\%, falling slightly below the 48.0\% R@1 of the top-performing member, AP-GeM (ResNet-101, LM18), even though the average R@1 of the ensemble members is 38.5\%. 
Consequently, applying the ensembling technique to these AP-GeM models diminishes the performance of the best-performing member.

\begin{figure}[t]
    \centering
    \includegraphics[width=0.7\columnwidth]{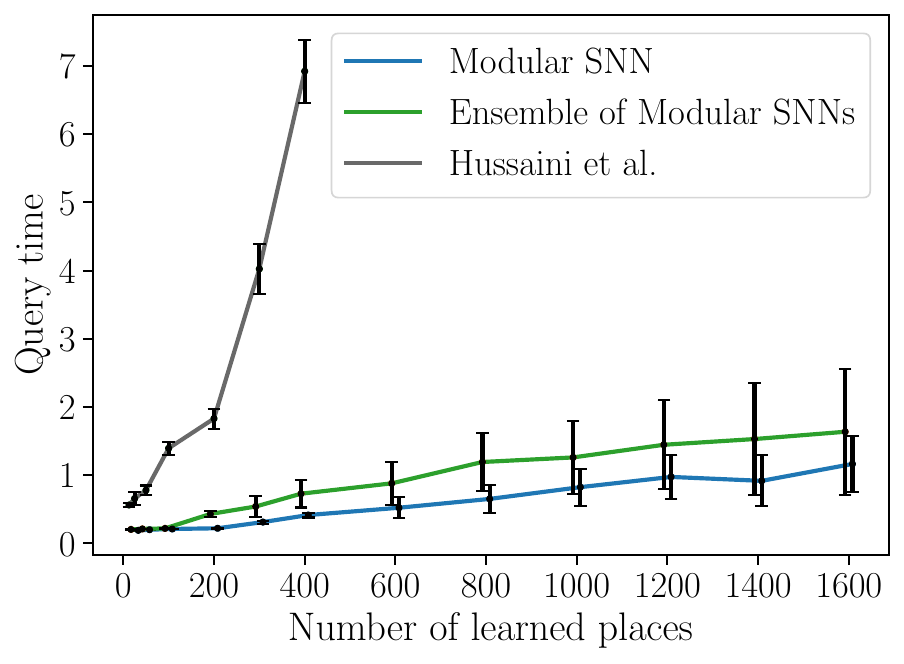}
    \vspace*{-0.25cm}
    \caption{\textbf{Performance scalability comparison:}
    We show the average query processing time for a single query image as the number of places (and thus overall network size) increases. We contrast our Modular and Ensemble of Modular SNNs with two ensemble members, with the approach from~\cite{Hussaini2022}. Both the Modular SNN and Ensemble of Modular SNNs demonstrate to scale linearly with the number of reference places. 
    In contrast, the Non-modular SNN did not scale well, and we were only able to test up to 6400 output neurons (400 places).
    We expect both our Ensemble and non-Ensemble variants to scale even better and yield even lower inference times on neuromorphic hardware due to their massive parallel processing capabilities.
    }
    \label{fig:scalability}
    \vspace*{-0.1cm}
\end{figure}

\subsection{Ablation Study on Member Randomization in Ensemble of Modular SNNs}
\label{res:abl_ModSNN_rand}

\begin{table}
    \caption{
    Ablation study on Ensemble of three Modular SNNs with and without randomization of initial weights, and shuffled order of images on the Oxford RobotCar dataset (Reference: Rain; query: Dusk). Best configuration is in \textbf{bold}.
    }
    \label{tab:abl_modSNN}
    \renewcommand{\arraystretch}{1.2}
    \centering
    \begin{tabular}{c|c|c}
        \textbf{Randomized Weights} & \textbf{Shuffled Order} & \textbf{R@1} \\ 
        \hline

        $\times$ & $\times$ & 0.10 \\
        $\checkmark$ & $\times$ & 0.11 \\
        $\times$ & $\checkmark$ &  0.19 \\
        $\checkmark$ & $\checkmark$ & \textbf{0.20} \\

    \end{tabular}
    \vspace{-0.2cm}
\end{table}

This section provides an ablation study on the R@1 performance of our Ensemble of Modular SNNs with and without input image order shuffling, and with and without different random weight initialization that is applied to the ensemble members. 
In our previous conference paper~\cite{hussaini2023ensembles}, we provided consecutive input images of a traverse to the modules for training. We initialized the weights of all modules using the same random values. 
Here, to increase the diversity among the ensemble members, we instead shuffled the reference images of each traverse for the training process, and then fed these shuffled images to the modules (\Cref{fig:frontpage_v2}). Moreover, we initialized the weights of each member, Modular SNN, using different random values, while within each member, using the same set of random values for all modules.
\Cref{tab:abl_modSNN} presents the R@1 performance of these four ensemble variants on the Oxford RobotCar dataset, where each ensemble contains three ensemble members. 
The first variant is an ensemble model where no randomization is applied, resulting in identical ensemble members and thus mirroring the performance of a single member, which is 10.4\%.
The ensemble members in the second variant differ only in their initial random weights, resulting in a R@1 of 11.0\%.
The third variant varies the ensemble members with only in the shuffled order of images, yielding in a R@1 performance of 19.1\%.
Lastly, the fourth variant combines both randomization of weights and shuffled order of images, producing the highest R@1 performance of 19.8\%.
Excluding the first variant, which negates the ensembling effect, the remaining three variants show similar R@1 performance improvements over their mean R@1 member performances. Randomizing the shuffled order of images significantly enhances ensemble performance compared to just randomizing the weights. The combination of both strategies obtains the highest R@1, albeit with a marginal improvement over the third variant, where only shuffled order of images are randomized.

\begin{figure*}[t]
  \centering
  \subfloat{
    \includegraphics[width=\linewidth]{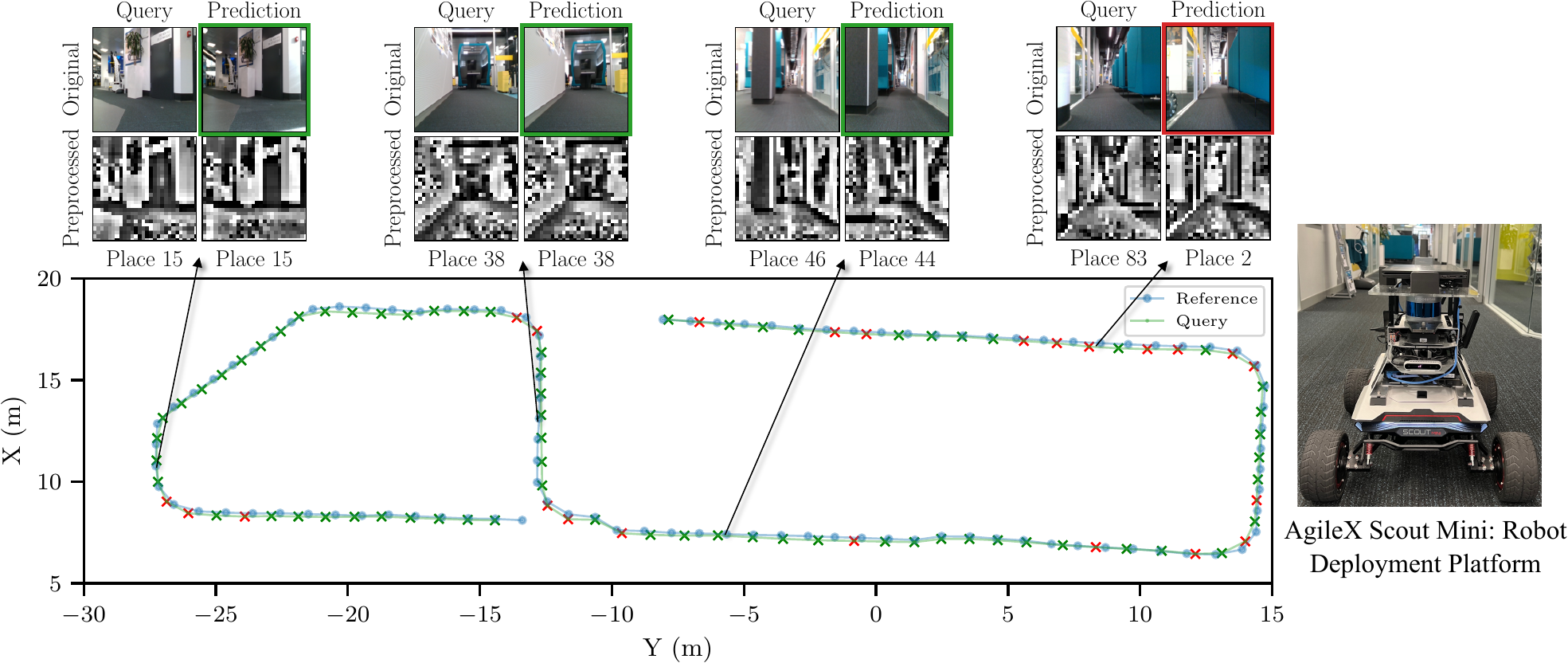}
    }
  \vspace*{-0.4cm}
  \caption{
  \textbf{Robot deployment feasibility study:}
  Left: Real-time deployment of our Modular SNN in a small indoor environment on a CPU. 
  First, the reference dataset is collected and the Modular SNN is trained offline. During inference, the robot moves through the environment, collecting images, and predicting their place labels in real-time. 
  The reference and query images are collected at different times of the day. The images were collected at 1 Hz with the robot moving at approximately 1 m/s. 
  The blue line and points represent the reference path and images, while the green line represents the query path. Correct predictions are marked with green crosses, and incorrect predictions are marked with red crosses. The four samples of the query and predicted places show the original images and their preprocessed forms, which are used as input to our Modular SNN.
  Right: Proof-of-concept robot deployment testing platform. We used an AgileX Scout Mini~\cite{agilex_scout_mini} robot equipped with an Intel RealSense D435 camera.
  }
  \label{fig:robot_testing_results}
  \vspace*{-0.2cm}
\end{figure*}

\subsection{Computational Scalability}
\label{res:compute_scalability}

\Cref{fig:scalability} provides insights into the scalability and computational efficiency of our Modular SNN and Ensemble of Modular SNNs with two ensemble members, against the Non-modular SNN from~\cite{Hussaini2022}. 
The plot shows that the query times of our Modular SNN and Ensemble of Modular SNNs scale linearly as the number of learned places increases. 
Meanwhile, the Non-modular SNN faces scalability issues beyond 400 places. 
We anticipate that implementing our Modular and Ensemble of Modular SNNs on neuromorphic hardware could substantially enhance processing speed through hardware parallelism. This is one of our future research directions, as described in more detail in~\Cref{discussions_conclusions}.

\subsection{Proof-of-Concept Robot Deployment}
\label{res:robot_deployment}

We also conducted a proof-of-concept deployment of our Modular SNN on AgileX's Scout Mini robot~\cite{agilex_scout_mini} in a small indoor environment operating in real-time on a CPU. 
In this experiment, we first collected  the reference dataset containing 100 images and trained our Modular SNN offline with four SNN modules (each assigned to learn 25 place labels). 
During inference, the robot moved through the environment, collecting images at 1 Hz, and predicting the place labels of the query images.
The robot’s speed was approximately 1 m/s during both the collection of the reference set and the inference time. 

\Cref{fig:robot_testing_results} shows the reference path taken by the robot, with correct and incorrect predictions of our Modular SNN at query time marked by green and red markers, respectively. 
The figure indicates that the model generally performs well, with a R@1 of 75.0\%. Most misclassifications occurred at curves or where the query image position is slightly shifted laterally and/or frontally. 
This is due to the limitation in viewpoint tolerance of our approach, which is discussed in~\Cref{discussions_conclusions} as part of our future work.

We used a ground truth tolerance of 3m, due to the high visual overlap between consecutive places.
The inference time of our Modular SNN ranged from 1.1 to 2.0 seconds per image.
This experiment validated the feasibility of our Modular SNN approach for real-time robot deployment in a small indoor environment.

\section{Conclusion}
\label{discussions_conclusions}

This paper has shed light on the capabilities of spiking neural networks (SNNs) in the realm of visual place recognition. Through a series of enhancements, we have showcased their utility and promise in this field.

Firstly, we introduced scalable SNNs that we dubbed Modular SNNs which represent a small region of the environment and have enhanced adaptability and efficiency in expansive environments. 
This innovation significantly broadens the applicability of SNNs for place recognition tasks.

Building on the foundation of Modular SNNs, we further enhanced our approach by introducing the Ensemble of Modular SNNs in our second contribution. In this case, multiple SNNs are employed to represent a single place, which demonstrated a substantial improvement in place recognition robustness and generalization ability. 
We have shown that the responsiveness of SNNs to ensembling is higher compared to conventional techniques that employed homogeneous and heterogeneous ensembles. This is evident as the average R@1 of our Ensemble of Modular SNNs across all datasets is consistently higher than the average R@1 of its ensemble members, highlighting that the ensembling technique significantly amplifies the capabilities of our Modular SNN approach. 
Moreover, our Ensemble of Modular SNNs has demonstrated competitive performance, in close proximity with some of the leading VPR methods, across various datasets.

Lastly, in addition to ensembling, we also explored the impact of sequence matching, a technique that further augments our system's performance by using multiple consecutive images for place matching. 
Pairing our Ensemble of Modular SNNs with sequence matching exhibited a higher R@1 performance improvement compared to VPR techniques with similar baselines, except in the case of the Oxford RobotCar dataset. This reinforces the significant role of sequence matching on enhancing SNN capabilities for visual place recognition.   
We also provided an indicator for sequence matching responsiveness, applicable to general VPR techniques, which demonstrated the competitive adaptability of our SNN-based solutions to sequence matching compared to that of VPR techniques.

Our work follows the similar trend seen in the recent state-of-the-art conventional VPR techniques such as~\cite{berton2022rethinking} that frame the visual place recognition problem as a classification task, enabling large-scale recognition capability by bypassing the computationally heavy process of computing the pairwise distance matrix for all query and reference feature representations. 
We highlight that our approach is trained only on the reference set, which has geographical overlap with the query set of the same dataset used for testing. This strategy is advantageous for real-world applications as it avoids pretraining on large datasets and uses fewer training images compared to conventional VPR techniques, which rely on extensive training datasets that are typically geographically separate from the reference and query set of the dataset used for evaluation.
Furthermore, we demonstrate the performance of our SNN approach using low resolution image sizes, which scales well with increasing number of places in terms of storage complexity.
In comparison, conventional VPR techniques typically need to store reference image feature descriptors, posing challenges for scaling to large datasets due to associated increase in storage requirements.

In the proof-of-concept CPU-based robot deployment of our work, we only used our Modular SNN approach, instead of our Ensemble of Modular SNN approach, because using multiple Modular SNNs to create the ensemble requires higher memory and results in an increase in latency. 
Future deployment of our SNN-based approach on neuromorphic hardware can provide significant improvements in the power usage and the latency of our approach.

Looking ahead, our future work aims to leverage these findings and explore new frontiers in neuromorphic computing. 
We plan to implement our approach on specialized neuromorphic hardware platforms, particularly Intel's Loihi 2~\cite{davies2021advancing}, to harness its inherent advantages in obtaining high energy efficiency and reduced latency. 
Although ensemble methods often face scalability issues, we see potential in neuromorphic computing, known for its exceptional parallel processing capabilities, to address these concerns. Such deployment will enable using our approach as a loop closure component for SLAM, in addition to using it as a re-localization method as presented in~\Cref{res:I}.
Implementing k-Nearest Neighbor (kNN) on Intel’s Loihi~\cite{davies2018loihi} has achieved comparable accuracy to CPU with 10 times less power and a latency of just 3.03 ms~\cite{frady2020neuromorphic}. 
A population-coded spiking network for robotic control on Loihi was 140 times more energy-efficient than on Jetson TX2, with similar performance~\cite{tang2021deep}. A spiking network for object classification on Loihi saw only a 3\% accuracy reduction compared to CPU, with 0.72 ms latency and 310 mW power consumption~\cite{viale2021carsnn}.

A limitation of our work is the minimal robustness to viewpoint shift, an aspect that most VPR techniques address effectively. 
Enhancing the resilience of our system to viewpoint change remains a priority for us, as it is crucial for reliable place recognition in more challenging situations. 
To overcome this challenge, we can incorporate an attention-based mechanism similar to recent transformer-based VPR approaches~\cite{keetha2023anyloc,wang2022transvpr}, or divide each input image into smaller patches, each processed by a module, similar to regional descriptors such as Patch-NetVLAD~\cite{hausler2021patch}. 
However, the necessity for robustness against significant viewpoint shifts may vary depending on the specific downstream application of our visual place recognition system. For instance, in scenarios where the VPR system acts as a loop closure component of a Simultaneous Localization and Mapping (SLAM) process, the limitations of the SLAM system in loop closure might render extreme viewpoint robustness less critical~\cite{tsintotas2022revisiting}.

We are also exploring the possibility of using event cameras~\cite{gallego2020event} to directly input event data to further enhance energy efficiency and reduce latency, moving away from our current strategy of converting traditional image data to rate-coded event streams. This includes adapting the components of our spiking network architecture, such as neuronal dynamics and learning mechanisms, to the sparse temporal nature of event data and modifying the definition of a place to suit event-based input characteristics.

Our research illuminates the significant potential of SNNs for robotic navigation, presenting a solution that is scalable, and robust for place recognition tasks. 
Our SNN-based approach is particularly responsive to ensembling and sequence matching techniques, as evidenced by its performance increase when these techniques are applied.
These techniques significantly enhance its robustness to high appearance changes, and its generalization ability across diverse environments.
These advancements in SNN technology not only enhance the efficiency of robotic navigation systems but also have vast applicability across various real-world robotics applications. Our findings are particularly promising for resource-constrained robots, such as those deployed in challenging environments including space and underwater, where the focus on edge computing and considerations for size, weight, and power emphasize its suitability for these rigorous settings.

\section{Acknowledgment}

The authors would like to thank the Queensland University of Technology (QUT) for continued support through the Centre for Robotics. The authors would also like to thank Dr. A. Hines,
T. Joseph, Dr. C. Malone, and G. B. Nair for their valuable insights on the drafts of this manuscript, and the QUT eResearch services for providing computational resources via the QUT High Performance Computing system.

\bibliographystyle{IEEEtran}
\bibliography{references}

\begin{thebibliography}{100}
\providecommand{\url}[1]{#1}
\csname url@rmstyle\endcsname
\providecommand{\newblock}{\relax}
\providecommand{\bibinfo}[2]{#2}
\providecommand\BIBentrySTDinterwordspacing{\spaceskip=0pt\relax}
\providecommand\BIBentryALTinterwordstretchfactor{4}
\providecommand\BIBentryALTinterwordspacing{\spaceskip=\fontdimen2\font plus
\BIBentryALTinterwordstretchfactor\fontdimen3\font minus
  \fontdimen4\font\relax}
\providecommand\BIBforeignlanguage[2]{{%
\expandafter\ifx\csname l@#1\endcsname\relax
\typeout{** WARNING: IEEEtran.bst: No hyphenation pattern has been}%
\typeout{** loaded for the language `#1'. Using the pattern for}%
\typeout{** the default language instead.}%
\else
\language=\csname l@#1\endcsname
\fi
#2}}

\bibitem{ghosh2009spiking}
S.~Ghosh-Dastidar and H.~Adeli, ``Spiking neural networks,'' \emph{Int. J.
  Neural Syst.}, vol.~19, no.~04, pp. 295--308, 2009.

\bibitem{sandamirskaya2022neuromorphic}
Y.~Sandamirskaya, M.~Kaboli, J.~Conradt, and T.~Celikel, ``Neuromorphic
  computing hardware and neural architectures for robotics,'' \emph{Sci.
  Robot.}, vol.~7, no.~67, p. eabl8419, 2022.

\bibitem{schuman2022opportunities}
C.~D. Schuman \emph{et~al.}, ``Opportunities for neuromorphic computing
  algorithms and applications,'' \emph{Nat. Comput. Sci.}, vol.~2, no.~1, pp.
  10--19, 2022.

\bibitem{yamazaki2022spiking}
K.~Yamazaki, V.-K. Vo-Ho, D.~Bulsara, and N.~Le, ``Spiking neural networks and
  their applications: A review,'' \emph{Brain Sci.}, vol.~12, no.~7, p. 863,
  2022.

\bibitem{gerstner2014neuronal}
W.~Gerstner, W.~M. Kistler, R.~Naud, and L.~Paninski, \emph{Neuronal dynamics:
  From single neurons to networks and models of cognition}.\hskip 1em plus
  0.5em minus 0.4em\relax Cambridge University Press, 2014.

\bibitem{nunes2022spiking}
J.~D. Nunes, M.~Carvalho, D.~Carneiro, and J.~S. Cardoso, ``Spiking neural
  networks: A survey,'' \emph{IEEE Access}, vol.~10, pp. 60\,738--60\,764,
  2022.

\bibitem{frady2020neuromorphic}
E.~P. Frady \emph{et~al.}, ``Neuromorphic nearest neighbor search using
  {Intel's Pohoiki Springs},'' in \emph{Proc. Neuro-inspired Comput. Elements
  Worksh.}, 2020.

\bibitem{davies2021advancing}
M.~Davies \emph{et~al.}, ``Advancing neuromorphic computing with {Loihi}: A
  survey of results and outlook,'' \emph{Proc. IEEE}, vol. 109, no.~5, pp.
  911--934, 2021.

\bibitem{pei2019towards}
J.~Pei, , \emph{et~al.}, ``Towards artificial general intelligence with hybrid
  {Tianjic} chip architecture,'' \emph{Nature}, vol. 572, no. 7767, pp.
  106--111, 2019.

\bibitem{furber2014spinnaker}
S.~B. Furber, F.~Galluppi, S.~Temple, and L.~A. Plana, ``The {SpiNNaker}
  project,'' \emph{Proc. IEEE}, vol. 102, no.~5, pp. 652--665, 2014.

\bibitem{yik2023neurobench}
J.~Yik, S.~H. Ahmed, \emph{et~al.}, ``{NeuroBench: Advancing} neuromorphic
  computing through collaborative, fair and representative benchmarking,''
  \emph{arXiv preprint arXiv:2304.04640}, 2023.

\bibitem{eshraghian2023training}
J.~K. Eshraghian \emph{et~al.}, ``Training spiking neural networks using
  lessons from deep learning,'' \emph{Proc. IEEE}, 2023.

\bibitem{schubert2023visual}
S.~Schubert, P.~Neubert, S.~Garg, M.~Milford, and T.~Fischer, ``Visual place
  recognition: A tutorial,'' \emph{IEEE Robotics \& Automation Magazine}, 2023.

\bibitem{garg2021your}
S.~Garg, T.~Fischer, and M.~Milford, ``Where is your place, visual place
  recognition?'' in \emph{Int. Jt. Conf. Artif. Intell.}, 2021, pp. 4416--4425.

\bibitem{Lowry2015}
S.~Lowry, N.~S{\"{u}}nderhauf, P.~Newman, J.~J. Leonard, D.~Cox, P.~Corke, and
  M.~J. Milford, ``{Visual place recognition: A survey},'' \emph{IEEE Trans.
  Robot.}, vol.~32, no.~1, pp. 1--19, 2015.

\bibitem{tsintotas2022visual}
K.~A. Tsintotas, L.~Bampis, and A.~Gasteratos, ``Visual place recognition for
  simultaneous localization and mapping,'' \emph{Autonomous Vehicles Volume 2:
  Smart Vehicles}, pp. 47--79, 2022.

\bibitem{masone2021survey}
C.~Masone and B.~Caputo, ``A survey on deep visual place recognition,''
  \emph{IEEE Access}, vol.~9, pp. 19\,516--19\,547, 2021.

\bibitem{zhang2021visual}
X.~Zhang, L.~Wang, and Y.~Su, ``Visual place recognition: A survey from deep
  learning perspective,'' \emph{Pattern Recognit.}, vol. 113, p. 107760, 2021.

\bibitem{cadena2016past}
C.~Cadena \emph{et~al.}, ``Past, present, and future of simultaneous
  localization and mapping: Toward the robust-perception age,'' \emph{IEEE
  Trans. Robot.}, vol.~32, no.~6, pp. 1309--1332, 2016.

\bibitem{tsintotas2022revisiting}
K.~A. Tsintotas, L.~Bampis, and A.~Gasteratos, ``The revisiting problem in
  simultaneous localization and mapping: A survey on visual loop closure
  detection,'' \emph{IEEE Trans. Intell. Transp. Syst.}, vol.~23, no.~11, pp.
  19\,929--19\,953, 2022.

\bibitem{frenkel2023bottom}
C.~Frenkel, D.~Bol, and G.~Indiveri, ``Bottom-up and top-down approaches for
  the design of neuromorphic processing systems: tradeoffs and synergies
  between natural and artificial intelligence,'' \emph{Proc. IEEE}, 2023.

\bibitem{auda1999modular}
G.~Auda and M.~Kamel, ``Modular neural networks: a survey,'' \emph{Int. J.
  Neural Syst.}, vol.~9, no.~02, pp. 129--151, 1999.

\bibitem{rauker2023toward}
T.~R{\"a}uker, A.~Ho, S.~Casper, and D.~Hadfield-Menell, ``Toward transparent
  {AI}: A survey on interpreting the inner structures of deep neural
  networks,'' in \emph{IEEE Conf. Secure Trustworthy Mach. Learn.}, 2023, pp.
  464--483.

\bibitem{amer2019review}
M.~Amer and T.~Maul, ``A review of modularization techniques in artificial
  neural networks,'' \emph{Artif. Intell. Rev.}, vol.~52, pp. 527--561, 2019.

\bibitem{colosi2020plug}
M.~Colosi \emph{et~al.}, ``Plug-and-play {SLAM}: A unified {SLAM} architecture
  for modularity and ease of use,'' in \emph{IEEE/RSJ Int. Conf. Intell. Robot.
  Syst.}, 2020, pp. 5051--5057.

\bibitem{dube2017segmatch}
R.~Dub{\'e} \emph{et~al.}, ``Segmatch: Segment based place recognition in {3D}
  point clouds,'' in \emph{IEEE Int. Conf. Robot. Autom.}, 2017, pp.
  5266--5272.

\bibitem{garg2017improving}
S.~Garg, A.~Jacobson, S.~Kumar, and M.~Milford, ``Improving condition-and
  environment-invariant place recognition with semantic place categorization,''
  in \emph{IEEE/RSJ Int. Conf. Intell. Robot. Syst.}, 2017, pp. 6863--6870.

\bibitem{blanco2019modular}
J.-L. Blanco-Claraco, ``A modular optimization framework for localization and
  mapping.'' in \emph{Robot. Sci. Syst.}, 2019.

\bibitem{ganaie2022ensemble}
M.~A. Ganaie, M.~Hu, A.~Malik, M.~Tanveer, and P.~Suganthan, ``Ensemble deep
  learning: A review,'' \emph{Eng. Appl. Artif. Intell.}, vol. 115, p. 105151,
  2022.

\bibitem{li2023deep}
W.~Li, Y.~Peng, M.~Zhang, L.~Ding, H.~Hu, and L.~Shen, ``Deep model fusion: A
  survey,'' \emph{arXiv preprint arXiv:2309.15698}, 2023.

\bibitem{yang2023survey}
Y.~Yang, H.~Lv, and N.~Chen, ``A survey on ensemble learning under the era of
  deep learning,'' \emph{Artif. Intell. Rev.}, vol.~56, no.~6, pp. 5545--5589,
  2023.

\bibitem{dietterich2000ensemble}
T.~G. Dietterich, ``Ensemble methods in machine learning,'' in \emph{Int.
  Worksh. Multiple Classifier Syst.}, 2000, pp. 1--15.

\bibitem{zhou2012ensemble}
Z.-H. Zhou, \emph{Ensemble methods: foundations and algorithms}.\hskip 1em plus
  0.5em minus 0.4em\relax CRC press, 2012.

\bibitem{sagi2018ensemble}
O.~Sagi and L.~Rokach, ``Ensemble learning: A survey,'' \emph{Wiley
  Interdiscip. Rev. Data Min. Knowl. Discov.}, vol.~8, no.~4, p. e1249, 2018.

\bibitem{wu2020eao}
Y.~Wu, Y.~Zhang, D.~Zhu, Y.~Feng, S.~Coleman, and D.~Kerr, ``{EAO-SLAM}:
  Monocular semi-dense object {SLAM} based on ensemble data association,'' in
  \emph{IEEE/RSJ Int. Conf. Intell. Robot. Syst.}, 2020, pp. 4966--4973.

\bibitem{procopio2009learning}
M.~J. Procopio, J.~Mulligan, and G.~Grudic, ``Learning terrain segmentation
  with classifier ensembles for autonomous robot navigation in unstructured
  environments,'' \emph{J. Field Robot.}, vol.~26, no.~2, pp. 145--175, 2009.

\bibitem{arcanjo2023music}
B.~Arcanjo \emph{et~al.}, ``A-music: An adaptive ensemble system for visual
  place recognition in changing environments,'' \emph{arXiv preprint
  arXiv:2303.14247}, 2023.

\bibitem{malone2023boosting}
C.~Malone, S.~Hausler, T.~Fischer, and M.~Milford, ``Boosting performance of a
  baseline visual place recognition technique by predicting the maximally
  complementary technique,'' in \emph{IEEE Int. Conf. Robot. Autom.}, 2023, pp.
  1919--1925.

\bibitem{fischer2020event}
T.~Fischer and M.~Milford, ``Event-based visual place recognition with
  ensembles of temporal windows,'' \emph{IEEE Robot. Autom. Lett.}, vol.~5,
  no.~4, pp. 6924--6931, 2020.

\bibitem{lakshminarayanan2017simple}
B.~Lakshminarayanan, A.~Pritzel, and C.~Blundell, ``Simple and scalable
  predictive uncertainty estimation using deep ensembles,'' \emph{Adv. Neural
  Inform. Process. Syst.}, vol.~30, 2017.

\bibitem{milford2012seqslam}
M.~J. Milford and G.~F. Wyeth, ``{SeqSLAM}: Visual route-based navigation for
  sunny summer days and stormy winter nights,'' in \emph{IEEE Int. Conf. Robot.
  Autom.}, 2012, pp. 1643--1649.

\bibitem{garg2021seqnet}
S.~Garg and M.~Milford, ``{SeqNet}: Learning descriptors for sequence-based
  hierarchical place recognition,'' \emph{IEEE Robot. Autom. Lett.}, vol.~6,
  no.~3, pp. 4305--4312, 2021.

\bibitem{garg2022seqmatchnet}
S.~Garg, M.~Vankadari, and M.~Milford, ``{SeqMatchNet}: Contrastive learning
  with sequence matching for place recognition \& relocalization,'' in
  \emph{Conference on Robot Learning}, 2022, pp. 429--443.

\bibitem{schubert2021fast}
S.~Schubert, P.~Neubert, and P.~Protzel, ``Fast and memory efficient graph
  optimization via icm for visual place recognition.'' in \emph{Robot. Sci.
  Syst.}, vol.~73, 2021.

\bibitem{mereu2022learning}
R.~Mereu, G.~Trivigno, G.~Berton, C.~Masone, and B.~Caputo, ``Learning
  sequential descriptors for sequence-based visual place recognition,''
  \emph{IEEE Robot. Autom. Lett.}, vol.~7, no.~4, pp. 10\,383--10\,390, 2022.

\bibitem{facil2019condition}
J.~M. Facil, D.~Olid, L.~Montesano, and J.~Civera, ``Condition-invariant
  multi-view place recognition,'' \emph{arXiv preprint arXiv:1902.09516}, 2019.

\bibitem{naseer2015robust}
T.~Naseer, M.~Ruhnke, C.~Stachniss, L.~Spinello, and W.~Burgard, ``Robust
  visual {SLAM} across seasons,'' in \emph{IEEE/RSJ Int. Conf. Intell. Robot.
  Syst.}, 2015, pp. 2529--2535.

\bibitem{torii201524}
A.~Torii, R.~Arandjelovic, J.~Sivic, M.~Okutomi, and T.~Pajdla, ``24/7 place
  recognition by view synthesis,'' in \emph{IEEE Conf. Comput. Vis. Pattern
  Recog.}, 2015, pp. 1808--1817.

\bibitem{Arandjelovic2018}
R.~Arandjelovic, P.~Gronat, A.~Torii, T.~Pajdla, and J.~Sivic, ``{NetVLAD: CNN}
  architecture for weakly supervised place recognition,'' \emph{IEEE Trans.
  Pattern Anal. Mach. Intell.}, vol.~40, no.~6, pp. 1437--1451, 2018.

\bibitem{revaud2019learning}
J.~Revaud, J.~Almaz{\'a}n, R.~S. Rezende, and C.~R.~d. Souza, ``Learning with
  average precision: Training image retrieval with a listwise loss,'' in
  \emph{Int. Conf. Comput. Vis.}, 2019, pp. 5107--5116.

\bibitem{leyva2023data}
M.~Leyva-Vallina, N.~Strisciuglio, and N.~Petkov, ``Data-efficient large scale
  place recognition with graded similarity supervision,'' in \emph{IEEE Conf.
  Comput. Vis. Pattern Recog.}, 2023, pp. 23\,487--23\,496.

\bibitem{berton2022rethinking}
G.~Berton, C.~Masone, and B.~Caputo, ``Rethinking visual geo-localization for
  large-scale applications,'' in \emph{IEEE Conf. Comput. Vis. Pattern Recog.},
  2022, pp. 4878--4888.

\bibitem{ali2023mixvpr}
A.~Ali-Bey, B.~Chaib-Draa, and P.~Giguere, ``Mixvpr: Feature mixing for visual
  place recognition,'' in \emph{IEEE/CVF Winter Conf. Appl. Comput. Vis.},
  2023, pp. 2998--3007.

\bibitem{sunderhauf2013we}
N.~S{\"u}nderhauf, P.~Neubert, and P.~Protzel, ``Are we there yet? {Challenging
  SeqSLAM} on a 3000 km journey across all four seasons,'' in \emph{IEEE Int.
  Conf. Robot. Autom. Worksh.}, 2013.

\bibitem{RobotCar}
W.~Maddern, G.~Pascoe, C.~Linegar, and P.~Newman, ``1 year, 1000 km: The
  {Oxford RobotCar} dataset,'' \emph{Int. J. Robot. Res.}, vol.~36, no.~1, pp.
  3--15, 2017.

\bibitem{bruce2015sfu}
J.~Bruce, J.~Wawerla, and R.~Vaughan, ``The {SFU} mountain dataset:
  Semi-structured woodland trails under changing environmental conditions,'' in
  \emph{IEEE Int. Conf. Robot. Autom.}, 2015.

\bibitem{ros2016synthia}
G.~Ros, L.~Sellart, J.~Materzynska, D.~Vazquez, and A.~M. Lopez, ``The synthia
  dataset: A large collection of synthetic images for semantic segmentation of
  urban scenes,'' in \emph{IEEE Conf. Comput. Vis. Pattern Recog.}, 2016, pp.
  3234--3243.

\bibitem{milford2008mapping}
M.~J. Milford and G.~F. Wyeth, ``Mapping a suburb with a single camera using a
  biologically inspired {SLAM} system,'' \emph{IEEE Trans. Robot.}, vol.~24,
  no.~5, pp. 1038--1053, 2008.

\bibitem{Hussaini2022}
S.~Hussaini, M.~Milford, and T.~Fischer, ``Spiking neural networks for visual
  place recognition via weighted neuronal assignments,'' \emph{IEEE Robot.
  Autom. Lett.}, vol.~7, no.~2, pp. 4094--4101, 2022.

\bibitem{agilex_scout_mini}
\BIBentryALTinterwordspacing
A.~Robotics, ``Scout mini: A small size 4wd mobile robot,'' 2024, accessed:
  2024-06-10. [Online]. Available:
  \url{https://global.agilex.ai/products/scout-mini}
\BIBentrySTDinterwordspacing

\bibitem{hussaini2023ensembles}
S.~Hussaini, M.~Milford, and T.~Fischer, ``Ensembles of compact,
  region-specific \& regularized spiking neural networks for scalable place
  recognition,'' in \emph{IEEE Int. Conf. Robot. Autom.}, 2023, pp. 4200--4207.

\bibitem{schuman2017survey}
C.~D. Schuman \emph{et~al.}, ``A survey of neuromorphic computing and neural
  networks in hardware,'' \emph{arXiv preprint arXiv:1705.06963}, 2017.

\bibitem{rueckauer2017conversion}
B.~Rueckauer, I.-A. Lungu, Y.~Hu, M.~Pfeiffer, and S.-C. Liu, ``Conversion of
  continuous-valued deep networks to efficient event-driven networks for image
  classification,'' \emph{Front. Neurosci.}, vol.~11, p. 682, 2017.

\bibitem{bu2021optimal}
T.~Bu, W.~Fang, J.~Ding, P.~DAI, Z.~Yu, and T.~Huang, ``Optimal {ANN-SNN}
  conversion for high-accuracy and ultra-low-latency spiking neural networks,''
  in \emph{Int. Conf. Learn. Represent.}, 2021.

\bibitem{ding2021optimal}
J.~Ding, Z.~Yu, Y.~Tian, and T.~Huang, ``Optimal {ANN-SNN} conversion for fast
  and accurate inference in deep spiking neural networks,'' in \emph{Int. Jt.
  Conf. Artif. Intell.}, 2021.

\bibitem{hunsberger2016training}
E.~Hunsberger and C.~Eliasmith, ``Training spiking deep networks for
  neuromorphic hardware,'' \emph{arXiv preprint arXiv:1611.05141}, 2016.

\bibitem{severa2019training}
W.~Severa, C.~M. Vineyard, R.~Dellana, S.~J. Verzi, and J.~B. Aimone,
  ``Training deep neural networks for binary communication with the whetstone
  method,'' \emph{Nat. Mach. Intell.}, vol.~1, no.~2, pp. 86--94, 2019.

\bibitem{stockl2021optimized}
C.~St{\"o}ckl and W.~Maass, ``Optimized spiking neurons can classify images
  with high accuracy through temporal coding with two spikes,'' \emph{Nat.
  Mach. Intell.}, vol.~3, no.~3, pp. 230--238, 2021.

\bibitem{lee2020enabling}
C.~Lee, S.~S. Sarwar, P.~Panda, G.~Srinivasan, and K.~Roy, ``Enabling
  spike-based backpropagation for training deep neural network architectures,''
  \emph{Front. Neurosci.}, p. 119, 2020.

\bibitem{renner2021backpropagation}
A.~Renner, F.~Sheldon, A.~Zlotnik, L.~Tao, and A.~Sornborger, ``The
  backpropagation algorithm implemented on spiking neuromorphic hardware,''
  \emph{arXiv preprint arXiv:2106.07030}, 2021.

\bibitem{shen2022backpropagation}
G.~Shen, D.~Zhao, and Y.~Zeng, ``Backpropagation with biologically plausible
  spatiotemporal adjustment for training deep spiking neural networks,''
  \emph{Patterns}, vol.~3, no.~6, 2022.

\bibitem{bi1998synaptic}
G.-q. Bi and M.-m. Poo, ``Synaptic modifications in cultured hippocampal
  neurons: dependence on spike timing, synaptic strength, and postsynaptic cell
  type,'' \emph{J. Neurosci.}, vol.~18, no.~24, pp. 10\,464--10\,472, 1998.

\bibitem{diehl2015unsupervised}
P.~U. Diehl and M.~Cook, ``Unsupervised learning of digit recognition using
  spike-timing-dependent plasticity,'' \emph{Front. Comput. Neurosci.}, vol.~9,
  no.~99, pp. 1--9, 2015.

\bibitem{abadia2021cerebellar}
I.~Abad{\'\i}a, F.~Naveros, E.~Ros, R.~R. Carrillo, and N.~R. Luque, ``A
  cerebellar-based solution to the nondeterministic time delay problem in
  robotic control,'' \emph{Sci. Robot.}, vol.~6, no.~58, p. eabf2756, 2021.

\bibitem{vitale2021event}
A.~Vitale, A.~Renner, C.~Nauer, D.~Scaramuzza, and Y.~Sandamirskaya,
  ``Event-driven vision and control for {UAVs} on a neuromorphic chip,'' in
  \emph{IEEE Int. Conf. Robot. Autom.}, 2021, pp. 103--109.

\bibitem{dupeyroux2021neuromorphic}
J.~Dupeyroux, J.~J. Hagenaars, F.~Paredes-Vall{\'e}s, and G.~C. de~Croon,
  ``Neuromorphic control for optic-flow-based landing of mavs using the loihi
  processor,'' in \emph{IEEE Int. Conf. Robot. Autom.}, 2021, pp. 96--102.

\bibitem{stagsted2020event}
R.~K. Stagsted \emph{et~al.}, ``Event-based {PID} controller fully realized in
  neuromorphic hardware: A one {DoF} study,'' in \emph{IEEE/RSJ Int. Conf.
  Intell. Robot. Syst.}, 2020, pp. 10\,939--10\,944.

\bibitem{ding2022biologically}
J.~Ding \emph{et~al.}, ``Biologically inspired dynamic thresholds for spiking
  neural networks,'' \emph{Adv. Neural Inform. Process. Syst.}, vol.~35, pp.
  6090--6103, 2022.

\bibitem{tieck2017towards}
J.~C.~V. Tieck \emph{et~al.}, ``Towards grasping with spiking neural networks
  for anthropomorphic robot hands,'' in \emph{IEEE Int. Conf. Artif. Neural
  Netw.}, 2017, pp. 43--51.

\bibitem{tieck2018controlling}
J.~C.~V. Tieck, L.~Steffen, J.~Kaiser, A.~Roennau, and R.~Dillmann,
  ``Controlling a robot arm for target reaching without planning using spiking
  neurons,'' in \emph{Int. Conf. Cogn. Inform. Cogn. Comput.}, 2018, pp.
  111--116.

\bibitem{oikonomou2023hybrid}
K.~M. Oikonomou, I.~Kansizoglou, and A.~Gasteratos, ``A hybrid spiking neural
  network reinforcement learning agent for energy-efficient object
  manipulation,'' \emph{Machines}, vol.~11, no.~2, p. 162, 2023.

\bibitem{lele2021end}
A.~Lele, Y.~Fang, J.~Ting, and A.~Raychowdhury, ``An end-to-end spiking neural
  network platform for edge robotics: From event-cameras to central pattern
  generation,'' \emph{IEEE Trans. Cogn. Develop. Syst.}, vol.~14, no.~3, pp.
  1092--1103, 2021.

\bibitem{luo2022conversion}
Y.~Luo, H.~Shen, X.~Cao, T.~Wang, Q.~Feng, and Z.~Tan, ``Conversion of siamese
  networks to spiking neural networks for energy-efficient object tracking,''
  \emph{Neural. Comput. Appl.}, vol.~34, no.~12, pp. 9967--9982, 2022.

\bibitem{kreiser2020chip}
R.~Kreiser, A.~Renner, V.~R. Leite, B.~Serhan, C.~Bartolozzi, A.~Glover, and
  Y.~Sandamirskaya, ``An on-chip spiking neural network for estimation of the
  head pose of the {iCub} robot,'' \emph{Front. Neurosci.}, vol.~14, p. 551,
  2020.

\bibitem{renner2022neuromorphic}
A.~Renner \emph{et~al.}, ``Neuromorphic visual scene understanding with
  resonator networks,'' \emph{arXiv preprint arXiv:2208.12880}, 2022.

\bibitem{galluppi2012live}
F.~Galluppi \emph{et~al.}, ``Live demo: Spiking {RatSLAM: Rat} hippocampus
  cells in spiking neural hardware,'' in \emph{IEEE Biomed. Circuits Syst.
  Conf.}, 2012, pp. 91--91.

\bibitem{tang2018gridbot}
G.~Tang and K.~P. Michmizos, ``Gridbot: An autonomous robot controlled by a
  spiking neural network mimicking the brain's navigational system,'' in
  \emph{Int. Conf. Neuromorphic Syst.}, 2018, pp. 1--8.

\bibitem{kreiser2018pose}
R.~Kreiser, A.~Renner, Y.~Sandamirskaya, and P.~Pienroj, ``Pose estimation and
  map formation with spiking neural networks: towards neuromorphic {SLAM},'' in
  \emph{IEEE/RSJ Int. Conf. Intell. Robot. Syst.}, 2018, pp. 2159--2166.

\bibitem{tang2019spiking}
G.~Tang, A.~Shah, and K.~P. Michmizos, ``Spiking neural network on neuromorphic
  hardware for energy-efficient unidimensional {SLAM},'' in \emph{IEEE/RSJ Int.
  Conf. Intell. Robot. Syst.}, 2019, pp. 4176--4181.

\bibitem{dumont2023exploiting}
N.~S.-Y. Dumont, P.~M. Furlong, J.~Orchard, and C.~Eliasmith, ``Exploiting
  semantic information in a spiking neural {SLAM} system,'' \emph{Front.
  Neurosci.}, vol.~17, 2023.

\bibitem{kreiser2018neuromorphic}
R.~Kreiser, M.~Cartiglia, J.~N. Martel, J.~Conradt, and Y.~Sandamirskaya, ``A
  neuromorphic approach to path integration: a head-direction spiking neural
  network with vision-driven reset,'' in \emph{IEEE Int. Symp. Circuits Syst.},
  2018, pp. 1--5.

\bibitem{kreiser2020error}
R.~Kreiser, G.~Waibel, N.~Armengol, A.~Renner, and Y.~Sandamirskaya, ``Error
  estimation and correction in a spiking neural network for map formation in
  neuromorphic hardware,'' in \emph{IEEE Int. Conf. Robot. Autom.}, 2020, pp.
  6134--6140.

\bibitem{safa2023fusing}
A.~Safa, T.~Verbelen, I.~Ocket, A.~Bourdoux, H.~Sahli, F.~Catthoor, and
  G.~Gielen, ``Fusing event-based camera and radar for {SLAM} using spiking
  neural networks with continual {STDP} learning,'' in \emph{IEEE Int. Conf.
  Robot. Autom.}, 2023, pp. 2782--2788.

\bibitem{milford2015place}
M.~Milford, H.~Kim, M.~Mangan, S.~Leutenegger, T.~Stone, B.~Webb, and
  A.~Davison, ``Place recognition with event-based cameras and a neural
  implementation of {SeqSLAM},'' \emph{arXiv preprint arXiv:1505.04548}, 2015.

\bibitem{weikersdorfer2013simultaneous}
D.~Weikersdorfer, R.~Hoffmann, and J.~Conradt, ``Simultaneous localization and
  mapping for event-based vision systems,'' in \emph{Int. Conf. Comput. Vis.
  Syst.}, 2013, pp. 133--142.

\bibitem{vidal2018ultimate}
A.~R. Vidal, H.~Rebecq, T.~Horstschaefer, and D.~Scaramuzza, ``Ultimate {SLAM?
  Combining} events, images, and imu for robust visual {SLAM} in {HDR} and
  high-speed scenarios,'' \emph{IEEE Robot. Autom. Lett.}, vol.~3, no.~2, pp.
  994--1001, 2018.

\bibitem{fischer2022many}
T.~Fischer and M.~Milford, ``How many events do you need? {Event-based} visual
  place recognition using sparse but varying pixels,'' \emph{IEEE Robot. Autom.
  Lett.}, vol.~7, no.~4, pp. 12\,275--12\,282, 2022.

\bibitem{gallego2020event}
G.~Gallego \emph{et~al.}, ``Event-based vision: A survey,'' \emph{IEEE Trans.
  Pattern Anal. Mach. Intell.}, vol.~44, no.~1, pp. 154--180, 2020.

\bibitem{xu2020probabilistic}
M.~Xu, N.~Snderhauf, and M.~Milford, ``Probabilistic visual place recognition
  for hierarchical localization,'' \emph{IEEE Robot. Autom. Lett.}, vol.~6,
  no.~2, pp. 311--318, 2020.

\bibitem{cummins2008fab}
M.~Cummins and P.~Newman, ``Fab-map: Probabilistic localization and mapping in
  the space of appearance,'' \emph{Int. J. Robot. Res.}, vol.~27, no.~6, pp.
  647--665, 2008.

\bibitem{doan2019scalable}
A.-D. Doan, Y.~Latif, T.-J. Chin, Y.~Liu, T.-T. Do, and I.~Reid, ``Scalable
  place recognition under appearance change for autonomous driving,'' in
  \emph{Int. Conf. Comput. Vis.}, 2019, pp. 9319--9328.

\bibitem{jegou2010aggregating}
H.~J{\'e}gou, M.~Douze, C.~Schmid, and P.~P{\'e}rez, ``Aggregating local
  descriptors into a compact image representation,'' in \emph{IEEE Conf.
  Comput. Vis. Pattern Recog.}, 2010, pp. 3304--3311.

\bibitem{trivigno2023divide}
G.~Trivigno, G.~Berton, J.~Aragon, B.~Caputo, and C.~Masone,
  ``Divide\&classify: Fine-grained classification for city-wide visual
  geo-localization,'' in \emph{Int. Conf. Comput. Vis.}, 2023, pp.
  11\,142--11\,152.

\bibitem{sarlin2019coarse}
P.-E. Sarlin, C.~Cadena, R.~Siegwart, and M.~Dymczyk, ``From coarse to fine:
  Robust hierarchical localization at large scale,'' in \emph{IEEE Conf.
  Comput. Vis. Pattern Recog.}, 2019, pp. 12\,716--12\,725.

\bibitem{fan2017biologically}
C.~Fan, Z.~Chen, A.~Jacobson, X.~Hu, and M.~Milford, ``Biologically-inspired
  visual place recognition with adaptive multiple scales,'' \emph{Robo. and
  Auton. sys.}, vol.~96, pp. 224--237, 2017.

\bibitem{hausler2020hierarchical}
S.~Hausler and M.~Milford, ``Hierarchical multi-process fusion for visual place
  recognition,'' in \emph{IEEE Int. Conf. Robot. Autom.}, 2020, pp. 3327--3333.

\bibitem{keetha2021hierarchical}
N.~V. Keetha, M.~Milford, and S.~Garg, ``A hierarchical dual model of
  environment-and place-specific utility for visual place recognition,''
  \emph{IEEE Robot. Autom. Lett.}, vol.~6, no.~4, pp. 6969--6976, 2021.

\bibitem{garcia2017hierarchical}
E.~Garcia-Fidalgo and A.~Ortiz, ``Hierarchical place recognition for
  topological mapping,'' \emph{IEEE Trans. Robot.}, vol.~33, no.~5, pp.
  1061--1074, 2017.

\bibitem{milford2004ratslam}
M.~J. Milford, G.~F. Wyeth, and D.~Prasser, ``{RatSLAM}: a hippocampal model
  for simultaneous localization and mapping,'' in \emph{IEEE Int. Conf. Robot.
  Autom.}, 2004, pp. 403--408.

\bibitem{neubert2019neurologically}
P.~Neubert, S.~Schubert, and P.~Protzel, ``A neurologically inspired sequence
  processing model for mobile robot place recognition,'' \emph{IEEE Robot.
  Autom. Lett.}, vol.~4, no.~4, pp. 3200--3207, 2019.

\bibitem{yu2019neuroslam}
F.~Yu, J.~Shang, Y.~Hu, and M.~Milford, ``{NeuroSLAM: A} brain-inspired {SLAM}
  system for {3D} environments,'' \emph{Biol. Cybern.}, vol. 113, no.~5, pp.
  515--545, 2019.

\bibitem{chancan2020hybrid}
M.~Chanc{\'a}n, L.~Hernandez-Nunez, A.~Narendra, A.~B. Barron, and M.~Milford,
  ``A hybrid compact neural architecture for visual place recognition,''
  \emph{IEEE Robot. Autom. Lett.}, vol.~5, no.~2, pp. 993--1000, 2020.

\bibitem{bing2023towards}
Z.~Bing, D.~Nitschke, G.~Zhuang, K.~Huang, and A.~Knoll, ``Towards cognitive
  navigation: A biologically inspired calibration mechanism for the head
  direction cell network,'' \emph{J. Artif. Intell.}, vol.~2, no.~1, pp.
  31--41, 2023.

\bibitem{tan2019evolving}
T.~Y. Tan, L.~Zhang, C.~P. Lim, B.~Fielding, Y.~Yu, and E.~Anderson, ``Evolving
  ensemble models for image segmentation using enhanced particle swarm
  optimization,'' \emph{IEEE access}, vol.~7, pp. 34\,004--34\,019, 2019.

\bibitem{fischer2018rt}
T.~Fischer, H.~J. Chang, and Y.~Demiris, ``{RT-GENE: Real-time} eye gaze
  estimation in natural environments,'' in \emph{Eur. Conf. Comput. Vis.},
  2018, pp. 334--352.

\bibitem{shim2016unsupervised}
Y.~Shim, A.~Philippides, K.~Staras, and P.~Husbands, ``Unsupervised learning in
  an ensemble of spiking neural networks mediated by {ITDP},'' \emph{PLoS
  Comput. Biol.}, vol.~12, no.~10, p. e1005137, 2016.

\bibitem{yang2022heterogeneous}
S.~Yang, B.~Linares-Barranco, and B.~Chen, ``Heterogeneous ensemble-based
  spike-driven few-shot online learning,'' \emph{Front. Neurosci.}, vol.~16, p.
  850932, 2022.

\bibitem{fu2021ensemble}
Q.~Fu and H.~Dong, ``An ensemble unsupervised spiking neural network for
  objective recognition,'' \emph{Neurocomputing}, vol. 419, pp. 47--58, 2021.

\bibitem{elbrecht2020evolving}
D.~Elbrecht \emph{et~al.}, ``Evolving ensembles of spiking neural networks for
  neuromorphic systems,'' in \emph{IEEE Symp. Ser. Comput. Intell.}, 2020, pp.
  1989--1994.

\bibitem{yin2012reservoir}
J.~Yin and Y.~Meng, ``Reservoir computing ensembles for multi-object behavior
  recognition,'' in \emph{Int. Jt. Conf. Neural Netw.}, 2012, pp. 1--8.

\bibitem{srinivasan2018spilinc}
G.~Srinivasan, P.~Panda, and K.~Roy, ``Spilinc: Spiking liquid-ensemble
  computing for unsupervised speech and image recognition,'' \emph{Front.
  Neurosci.}, vol.~12, p. 524, 2018.

\bibitem{panda2017ensemblesnn}
P.~Panda, G.~Srinivasan, and K.~Roy, ``{EnsembleSNN: Distributed} assistive
  {STDP} learning for energy-efficient recognition in spiking neural
  networks,'' in \emph{Int. Joint Conf. Neural Networks}, 2017, pp. 2629--2635.

\bibitem{naseer2014robust}
T.~Naseer, L.~Spinello, W.~Burgard, and C.~Stachniss, ``Robust visual robot
  localization across seasons using network flows,'' in \emph{AAAI Conf. Artif.
  Intell.}, vol.~28, no.~1, 2014.

\bibitem{hansen2014visual}
P.~Hansen and B.~Browning, ``Visual place recognition using hmm sequence
  matching,'' in \emph{IEEE/RSJ Int. Conf. Intell. Robot. Syst.}, 2014, pp.
  4549--4555.

\bibitem{arroyo2015towards}
R.~Arroyo, P.~F. Alcantarilla, L.~M. Bergasa, and E.~Romera, ``Towards
  life-long visual localization using an efficient matching of binary sequences
  from images,'' in \emph{IEEE Int. Conf. Robot. Autom.}, 2015, pp. 6328--6335.

\bibitem{garg2020delta}
S.~Garg, B.~Harwood, G.~Anand, and M.~Milford, ``Delta descriptors:
  Change-based place representation for robust visual localization,''
  \emph{IEEE Robot. Autom. Lett.}, vol.~5, no.~4, pp. 5120--5127, 2020.

\bibitem{xu2022deep}
M.~Xu, S.~Garg, M.~Milford, and S.~Gould, ``Deep declarative dynamic time
  warping for end-to-end learning of alignment paths,'' in \emph{Int. Conf.
  Learn. Represent.}, 2022.

\bibitem{hausler2021patch}
S.~Hausler, S.~Garg, M.~Xu, M.~Milford, and T.~Fischer, ``{Patch-NetVLAD:
  Multi-scale} fusion of locally-global descriptors for place recognition,'' in
  \emph{IEEE Conf. Comput. Vis. Pattern Recog.}, 2021, pp. 14\,141--14\,152.

\bibitem{camara2020visual}
L.~G. Camara and L.~P{\v{r}}eu{\v{c}}il, ``Visual place recognition by spatial
  matching of high-level {CNN} features,'' \emph{Rob. Auton. Syst.}, vol. 133,
  p. 103625, 2020.

\bibitem{hausler2019multi}
S.~Hausler, A.~Jacobson, and M.~Milford, ``Multi-process fusion: Visual place
  recognition using multiple image processing methods,'' \emph{IEEE Robot.
  Autom. Lett.}, vol.~4, no.~2, pp. 1924--1931, 2019.

\bibitem{molloy2020intelligent}
T.~L. Molloy, T.~Fischer, M.~Milford, and G.~N. Nair, ``Intelligent reference
  curation for visual place recognition via bayesian selective fusion,''
  \emph{IEEE Robot. Autom. Lett.}, vol.~6, no.~2, pp. 588--595, 2020.

\bibitem{neubert2021hyperdimensional}
P.~Neubert and S.~Schubert, ``Hyperdimensional computing as a framework for
  systematic aggregation of image descriptors,'' in \emph{IEEE Conf. Comput.
  Vis. Pattern Recog.}, 2021, pp. 16\,938--16\,947.

\bibitem{lowry2018lightweight}
S.~Lowry and H.~Andreasson, ``Lightweight, viewpoint-invariant visual place
  recognition in changing environments,'' \emph{IEEE Robot. Autom. Lett.},
  vol.~3, no.~2, pp. 957--964, 2018.

\bibitem{zaffar2021vpr}
M.~Zaffar \emph{et~al.}, ``{VPR-bench: An} open-source visual place recognition
  evaluation framework with quantifiable viewpoint and appearance change,''
  \emph{Int. J. Comput. Vis.}, vol. 129, no.~7, pp. 2136--2174, 2021.

\bibitem{simonyan2014very}
K.~Simonyan and A.~Zisserman, ``Very deep convolutional networks for
  large-scale image recognition,'' \emph{arXiv preprint arXiv:1409.1556}, 2014.

\bibitem{russakovsky2015imagenet}
O.~Russakovsky \emph{et~al.}, ``Imagenet large scale visual recognition
  challenge,'' \emph{Int. J. Comput. Vis}, vol. 115, pp. 211--252, 2015.

\bibitem{noh2017large}
H.~Noh, A.~Araujo, J.~Sim, T.~Weyand, and B.~Han, ``Large-scale image retrieval
  with attentive deep local features,'' in \emph{Int. Conf. Comput. Vis.},
  2017, pp. 3456--3465.

\bibitem{warburg2020mapillary}
F.~Warburg \emph{et~al.}, ``Mapillary street-level sequences: A dataset for
  lifelong place recognition,'' in \emph{IEEE Conf. Comput. Vis. Pattern
  Recog.}, 2020, pp. 2626--2635.

\bibitem{torii2013visual}
A.~Torii, J.~Sivic, T.~Pajdla, and M.~Okutomi, ``Visual place recognition with
  repetitive structures,'' in \emph{IEEE Conf. Comput. Vis. Pattern Recog.},
  2013, pp. 883--890.

\bibitem{lowe2004distinctive}
D.~G. Lowe, ``Distinctive image features from scale-invariant keypoints,''
  \emph{Int. J. Comput. Vis}, vol.~60, pp. 91--110, 2004.

\bibitem{radenovic2018fine}
F.~Radenovi{\'c}, G.~Tolias, and O.~Chum, ``Fine-tuning {CNN} image retrieval
  with no human annotation,'' \emph{IEEE Trans. Pattern Anal. Mach. Intell.},
  vol.~41, no.~7, pp. 1655--1668, 2018.

\bibitem{he2016deep}
K.~He, X.~Zhang, S.~Ren, and J.~Sun, ``Deep residual learning for image
  recognition,'' in \emph{IEEE Conf. Comput. Vis. Pattern Recog.}, 2016, pp.
  770--778.

\bibitem{gordo2016deep}
A.~Gordo, J.~Almaz{\'a}n, J.~Revaud, and D.~Larlus, ``Deep image retrieval:
  Learning global representations for image search,'' in \emph{Eur. Conf.
  Comput. Vis.}, 2016, pp. 241--257.

\bibitem{wang2018cosface}
H.~Wang \emph{et~al.}, ``Cosface: Large margin cosine loss for deep face
  recognition,'' in \emph{IEEE Conf. Comput. Vis. Pattern Recog.}, 2018, pp.
  5265--5274.

\bibitem{schubert2021makes}
S.~Schubert and P.~Neubert, ``What makes visual place recognition easy or
  hard?'' \emph{arXiv:2106.12671}, 2021.

\bibitem{davies2018loihi}
M.~Davies \emph{et~al.}, ``Loihi: A neuromorphic manycore processor with
  on-chip learning,'' \emph{IEEE Micro}, vol.~38, no.~1, pp. 82--99, 2018.

\bibitem{tang2021deep}
G.~Tang, N.~Kumar, R.~Yoo, and K.~Michmizos, ``Deep reinforcement learning with
  population-coded spiking neural network for continuous control,'' in
  \emph{Conference on Robot Learning}, 2021, pp. 2016--2029.

\bibitem{viale2021carsnn}
A.~Viale \emph{et~al.}, ``Carsnn: An efficient spiking neural network for
  event-based autonomous cars on the loihi neuromorphic research processor,''
  in \emph{Int. Jt. Conf. Neural Netw.}, 2021, pp. 1--10.

\bibitem{keetha2023anyloc}
N.~Keetha \emph{et~al.}, ``Anyloc: Towards universal visual place
  recognition,'' \emph{IEEE Robot. Autom. Lett.}, 2023.

\bibitem{wang2022transvpr}
R.~Wang \emph{et~al.}, ``Transvpr: Transformer-based place recognition with
  multi-level attention aggregation,'' in \emph{IEEE Conf. Comput. Vis. Pattern
  Recog.}, 2022, pp. 13\,648--13\,657.

\end{thebibliography}

\begin{IEEEbiography}[{\includegraphics[width=1in,height=1.25in,clip,keepaspectratio]{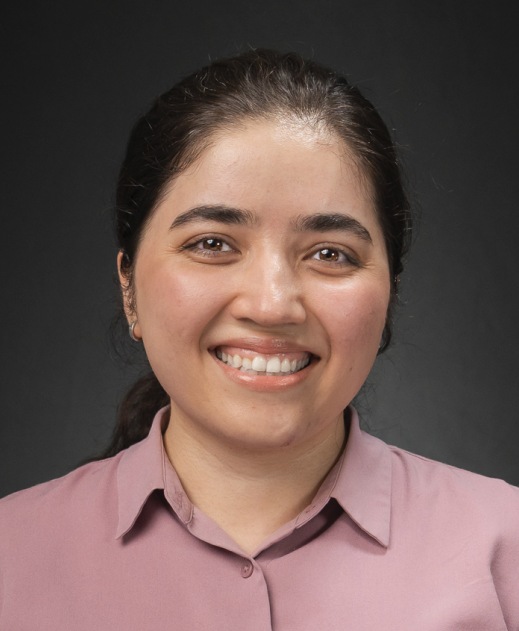}}]{Somayeh Hussaini} (Member, IEEE) received the Bachelor of Engineering degree in mechatronics with first class honours in 2020 from the Queensland University of Technology (QUT), Brisbane, QLD, Australia,
where she is currently working toward the Ph.D. degree in robotics, titled ``Spiking Neural Networks for Scalable Visual Place Recognition'', since 2021.
In 2024, she started her role as a Postdoctoral Research Fellow at QUT. Her research interests include robotics, computer vision, and neuromorphic computing. 
\end{IEEEbiography}

\begin{IEEEbiography}[{\includegraphics[width=1in,height=1in,clip,keepaspectratio]{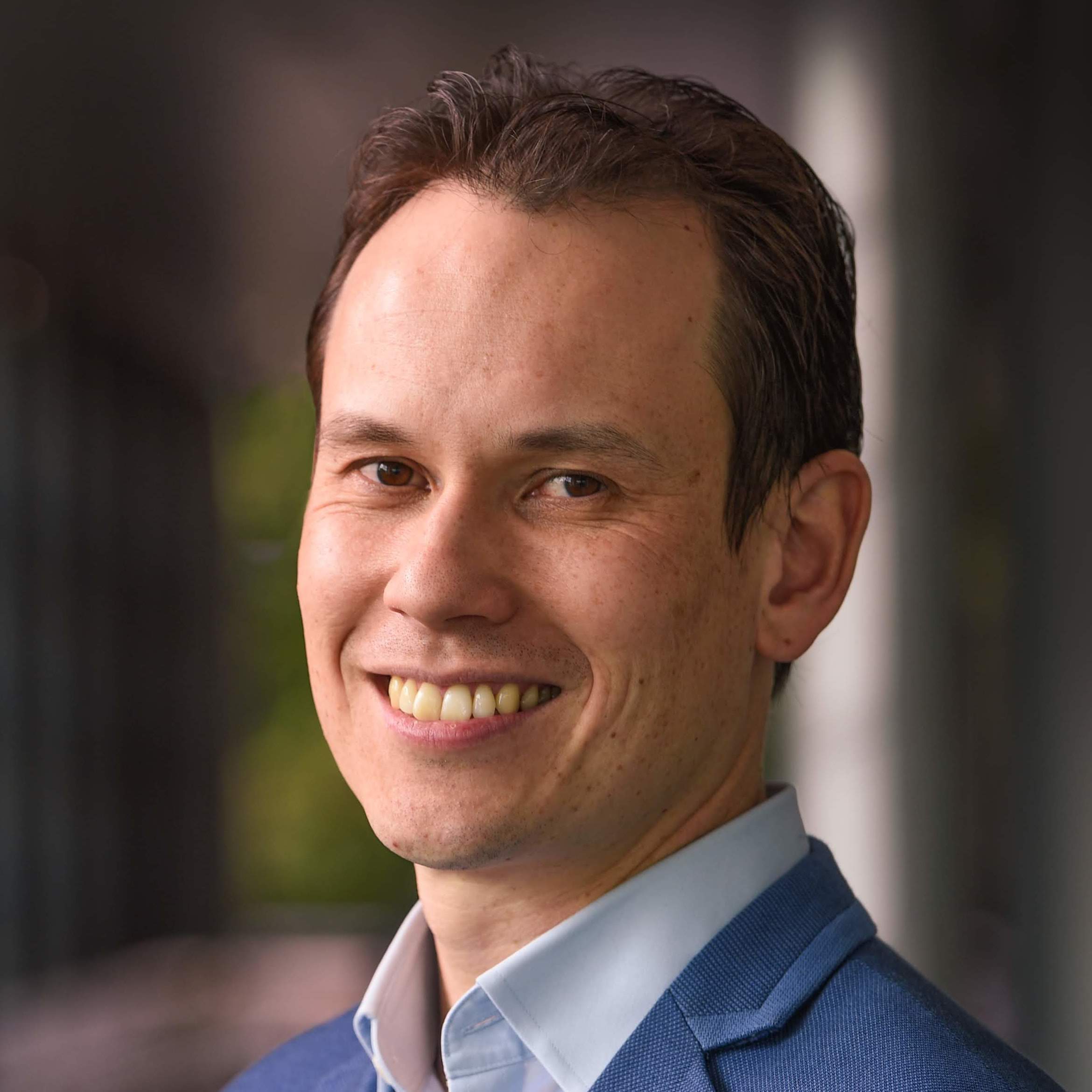}}]{Michael Milford, FTSE} (Senior Member, IEEE) 
received the bachelor degree in mechanical and space engineering and the Ph.D. degree in electrical engineering from The University of Queensland, Brisbane, QLD, Australia, in 2002 and 2006, respectively. 

He is currently the Director with the QUT Centre for Robotics, a Professor with the Queensland University of Technology, Brisbane, and is a Microsoft Research Faculty Fellow. His research interests include the neural mechanisms in the brain underlying tasks such as navigation and perception to develop new technologies in challenging application domains such as all-weather, anytime positioning for autonomous vehicles. 

Dr. Milford is a Fellow of the Australian Academy of Technology and Engineering and an Australian Research Council Laureate Fellow.

\end{IEEEbiography}

\begin{IEEEbiography}[{\includegraphics[width=1in,height=1.25in,clip,keepaspectratio]{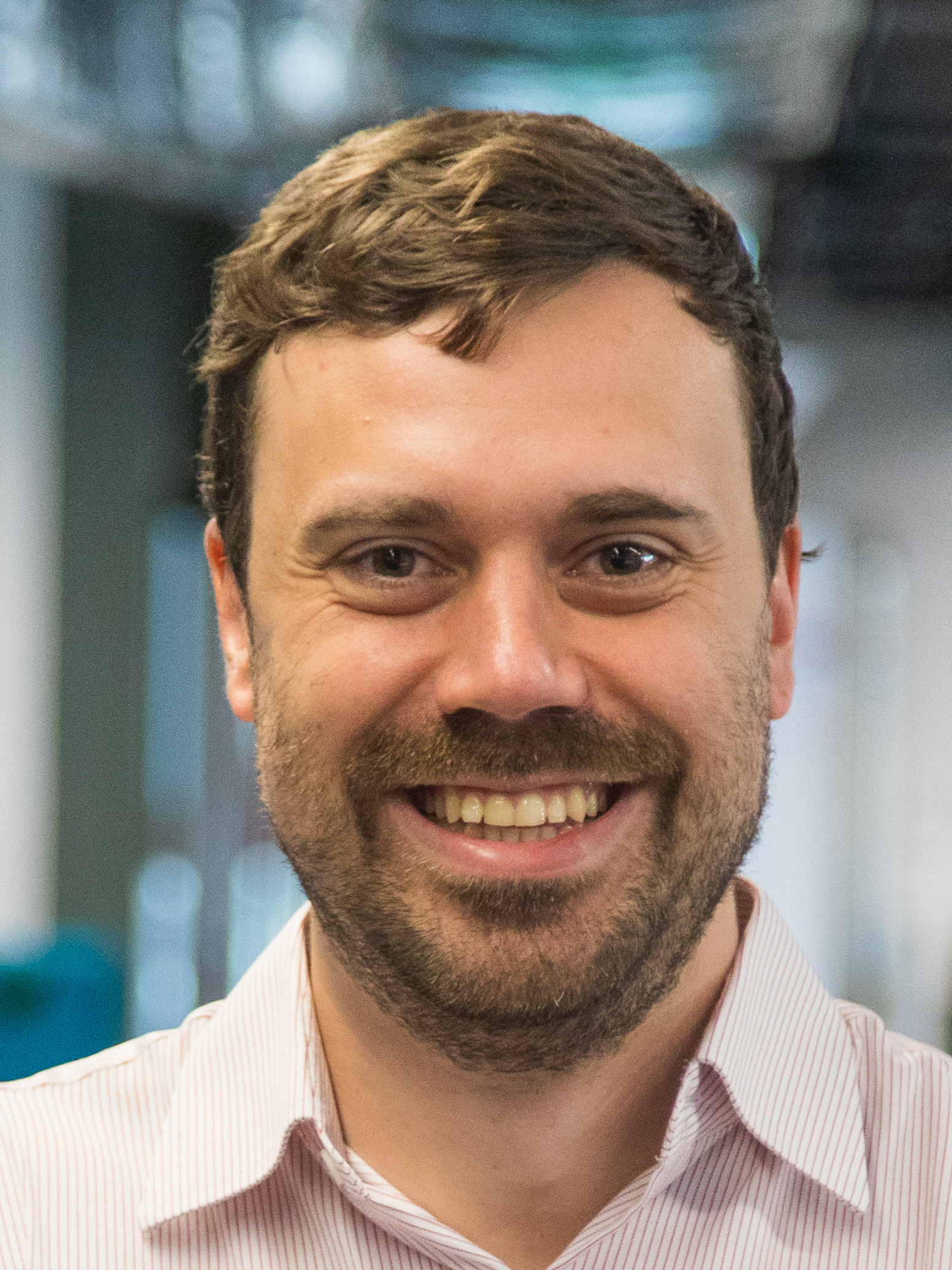}}]{Tobias Fischer} (Senior Member, IEEE) 
received the B.Sc. degree in computer engineering from the Ilmenau University of Technology, Ilmenau, Germany, in 2013, the M.Sc. degree in artificial intelligence from the University of Edinburgh, Edinburgh, U.K., in 2014, and the Ph.D. degree in robotics from the Personal Robotics Laboratory, Imperial College London, London, U.K., in 2018.

He combines his expertise in robotics, computer vision, and artificial intelligence to provide robots with perceptual abilities allowing safe, intelligent interactions with humans in real-world environments.

Dr. Fischer was the recipient of the prestigious Discovery Early Career Researcher Award (DECRA) by the Australian Research Council. His Ph.D. thesis received the U.K. Best Thesis in Robotics Award 2018 and the Eryl Cadwaladr Davies Award for the best thesis in Imperial’s Electrical and Electronic Engineering Department in 2017–2018. He was also the recipient of multiple best paper awards, including the 2023 IEEE TRANSACTIONS ON COGNITIVE AND DEVELOPMENTAL SYSTEMS OUTSTANDING PAPER AWARD.

\end{IEEEbiography}

\end{document}